\documentclass{article}

\usepackage{arxiv}

\usepackage[utf8]{inputenc} 
\usepackage[T1]{fontenc}    
\usepackage{hyperref}       
\usepackage{url}            
\usepackage{booktabs}       
\usepackage{amsfonts}       
\usepackage{nicefrac}       
\usepackage{microtype}      
\usepackage{lipsum}
\usepackage{fancyhdr}       
\usepackage{graphicx}       
\graphicspath{{media/}}     
\usepackage{color}
\usepackage{amsmath}
\usepackage{array}
\usepackage{tabularx}
\usepackage{colortbl}

\title{SuperPoint-E: local features for 3D reconstruction via tracking adaptation in endoscopy}

\author{
  O. L. Barbed, J.M. Martínez Montiel, A. C. Murillo \\
  University of Zaragoza \\
  Zaragoza, Spain\\
  leon@unizar.es\\
   \And
  P. Fua \\
  École Polytechnique Fédérale de Lausanne \\
  Lausanne, Switzerland\\
}

\begin{document}
\maketitle

\begin{abstract}
In this work, we focus on boosting the feature extraction to improve the performance of Structure-from-Motion (SfM) in endoscopy videos. We present SuperPoint-E, a new local feature extraction method that, using our proposed Tracking Adaptation supervision strategy, significantly improves the quality of feature detection and description in endoscopy. 
Extensive experimentation on real endoscopy recordings studies our approach’s most suitable configuration and evaluates SuperPoint-E feature quality. The comparison with other baselines also shows that our 3D reconstructions are denser and cover more and longer video segments because our detector fires more densely and our features are more likely to survive (i.e. higher detection precision). In addition, our descriptor is more discriminative, making the guided matching step almost redundant.
The presented approach brings significant improvements in the 3D reconstructions obtained, via SfM on endoscopy videos, compared to the original SuperPoint and the gold standard SfM COLMAP pipeline.
\end{abstract}


\section{Introduction}\label{sec1}
\label{sec:introduction}
Endoscopic examination involves inserting a light and camera attached to the end of a flexible tube into the body's hollow organs and cavities for exploration purposes. Its primary purpose is to assess the organs' condition and to diagnose, monitor, and treat various medical conditions. Incorporating image 3D reconstruction techniques into the workflow would allow for automated coverage assessment and for the use of mixed reality tools to assist the physician during the procedures. These additions to regular practice can significantly improve the procedures' quality and exploration times.

Widespread 3D reconstruction techniques, mostly feature based  approaches for Structure from Motion (SfM) and Simultaneous Location And Mapping (SLAM), have been tried for endoscopic 3D reconstruction~\cite{widya2019whole,ma2021rnnslam,rodriguez2024nr}.
However, they have only achieved limited success 
for several reasons. The cavities being explored often lack texture, which makes finding exploitable feature points, required by most of these methods, challenging. 
Water and other liquids present inside the organs very frequently reflect the light and form specularities, which partially obstruct the visibility while giving no 3D information, so they are problematic and are often ignored~\cite{barbed2022superpoint}.
Moreover, since the light is attached to a moving camera, the illumination can change drastically from one frame to the next. 
These challenges negatively affect the outcome of the initial steps of feature based SfM/SLAM, namely local feature  extraction (including point detection and descriptor computation). 

The main contribution in this work is \mbox{SuperPoint-E}, extending SuperPoint~\cite{detone2018superpoint}, a seminal method for deep learning based feature detection and description that 
is still inspiring recent approaches for keypoint detection~\cite{barroso2022key}, also being a competitive baseline in the performance comparisons.
Our proposed approach demonstrates superior domain adaptation to endoscopy imagery, and extracts feature points that are repeatable, reliable and discriminative. 
The benefits of the presented feature extraction approach are demonstrated in an SfM application on real endoscopy sequences. 
Note that an improvement in feature adequacy for SfM generalizes to SLAM, but SfM simplifies experimentation without hard constraints on real-time execution.

The key novelty in \mbox{SuperPoint-E} is our proposed training strategy, \textit{Tracking Adaptation}, that obtains groundtruth supervision from endoscopy SfM. An overview of this process is shown in Fig.~\ref{fig:method}. COLMAP~\cite{schoenberger2016sfm} reconstructions provide examples of reliable feature points, i.e., those that have been reconstructed. These points are used to supervise the training process. 
This work extends our preliminary SuperPoint-E version described in a conference paper~\cite{barbed2023tracking}. 
We present a refined version of our preliminary method with a more thorough experimentation and validation. 
New and diverse experiments are performed with additional data, including results with complete colonoscopy procedure recordings, to study the coverage of the reconstructions, and data from other domains, gastroscopy and bronchoscopy,  to illustrate the generalization capabilities. 
A more thorough analysis of the approach and its configuration, leading to an optimized setup, significantly improves model performance. 
The validation is more comprehensive thanks to new relevant metrics for the feature quality analysis, related to detection precision and reconstruction density.  



Our results demonstrate that \mbox{SuperPoint-E} provides significantly better features for the endoscopic domain compared to SuperPoint and SIFT. We compare these feature extractors on the gold standard SfM  COLMAP pipeline. \mbox{SuperPoint-E} 
is able to detect many more feature points that are well-suited for matching and triangulation, with a higher percentage of them being reconstructed when used for SfM. 
The descriptors of these feature points are more discriminative, 
making the guided matching step almost redundant. 
Besides, the 3D reconstructions obtained are much more dense and cover more and longer parts of the complete endoscopic videos used in the validation~\cite{azagra2023endomapper}.

\section{Related Work}\label{sec:relatedwork}

{\bf 3D reconstruction in endoscopy.}
3D reconstruction remains an open problem for medical settings such as laparoscopy and endoscopy. 
These applications are of high interest for the scientific community, and several endoscopic datasets have been published for this or similar purposes, e.g., HyperKvasir~\cite{borgli2020hyperkvasir}, EndoSLAM~\cite{ozyoruk2021endoslam} and EndoMapper~\cite{azagra2023endomapper}. 
With the popularization of Neural Radiance Field (NeRF) methods,  
several works have applied them to the endoscopic domain. Zha \textit{et al}.~\cite{zha2023endosurf} apply a SDF formulation to model tissue in stereo endoscopy sequences. Batlle \textit{et al}.~\cite{batlle2023lightneus} exploit the relation between depth and illumination decay to improve the representations in phantom sequences.

Earlier works like Grasa \textit{et al}.~\cite{grasa2013visual} have evaluated the performance of modern SLAM approaches on en\-dos\-co\-pic sequences. Mahmoud \textit{et al}.~\cite{mahmoud2018live} improved the performance of such methods in la\-pa\-ros\-co\-pic sequences. Other approaches take advantage of depth estimation neural networks to create dense SLAM re\-cons\-truc\-tions for surface coverage evaluation~\cite{ma2021rnnslam}. Liu \textit{et al.}~\cite{liu2022sage} focus on improving the reconstructions learning simultaneously appearance and geometry of the scene. Some recent approaches attempt to tackle specific en\-dos\-co\-py challenges, such as the deformation~\cite{rodriguez2022tracking} or the artifacts due to spe\-cu\-lar reflections in feature point detection~\cite{barbed2022superpoint}. Liu \textit{et al}.~\cite{liu2020extremely} is the closest approach to ours. They propose  ground\-truth supervision from 3D SfM reconstructions to train a dense description network for en\-dos\-co\-py image. We propose to further leverage the ground\-truth from 3D SfM reconstructions based on different feature detectors to train a network for not only description but also detection. 
Besides, our supervision comes from 
tracks of reconstructed points rather than pairs of corresponding points, encouraging discriminative and repeatable features.\\

\noindent
{\bf Local feature extraction.} 
Well known SfM and SLAM pipelines rely on accurate and robust local feature extraction methods. 
COLMAP~\cite{schoenberger2016sfm}, 
a public SfM tool, uses SIFT
while ORB-SLAM~\cite{mur2015orb} extracts ORB
features because of their efficiency. Both these feature extraction methods count with classical, hand-crafted descriptors that allowed to build such complex applications. However, transferring that performance to endoscopy settings remains a difficult task due to a wide range of challenges. Illumination artifacts, motion blur or the lack of texture result in low amount of correspondences along real endoscopy videos, which motivates the need for improved feature extraction strategies for this specific domain.

Research in deep learning methods for feature extraction and matching has been a very active field in the recent decade. The survey by Ma \textit{et al}.~\cite{ma2020image} presents a comprehensive introduction of deep learning methods to feature detection and matching. Our focus is on the feature detection and description stages. 
Early improvements in feature description were achieved by advanced training losses such as the triplet loss~\cite{mishchuk2017working}, and also choosing even more negative samples~\cite{tian2017l2}. However, approaches like LIFT~\cite{yi2016lift} opened the door for learning-based methods that directly replace classical feature extraction methods in the image registration pipeline. In this  line, SuperPoint~\cite{detone2018superpoint} proposes a self-supervised approach that obtains supervision via \textit{homographic adaptation}. 
Our work improves the performance of SuperPoint~\cite{detone2018superpoint} on endoscopy images. We base our work on SuperPoint because it is a seminal work that has inspired many follow up works, and is still among the top performers on current feature matching challenges~\cite{jin2021image}. Moreover, SuperPoint features are still being used as backbone for recent state-of-the-art feature matching methods~\cite{lindenberger2023lightglue}. 
Similar to DeTone \textit{et al.}~\cite{detone2018self}, we consider  improvements on feature extraction that provide good properties for downstream tasks. They design an end-to-end method to optimize the visual odometry computed with their features. Differently, 
we propose to supervise our training with points successfully used for 3D reconstruction by existing SfM pipelines. With this supervision, we train a model to extract more features of better quality, with good properties for SfM algorithms in endoscopy, e.g., being spread and out of large specularities.

\section{Tracking Adaptation for Local Feature Learning}
\label{sec:methods}

\begin{figure*}
    \centering
    \begin{tabular}{@{}c@{}}
         \includegraphics[width=0.8\linewidth]{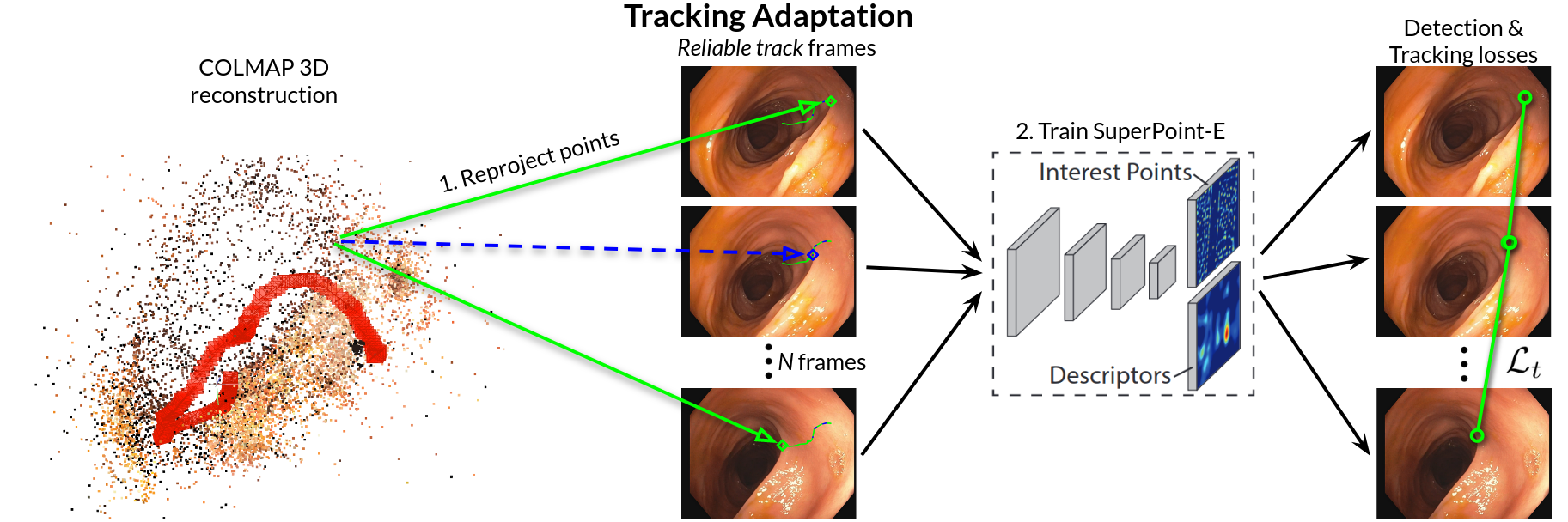} \\
    \end{tabular}
    \caption{\textit{Tracking Adaptation} overview. We run COLMAP 3D reconstructions and reproject the 3D points onto the sequence frames. Resulting \textit{reliable tracks} are used as supervision to train  \mbox{SuperPoint-E}. A new tracking loss leverages this supervision to refine the feature descriptors.}
    \label{fig:method}
\end{figure*}

SuperPoint~\cite{detone2018superpoint} is trained with their  \textit{Homographic Adaptation} strategy. We propose the alternative \textbf{\textit{Tracking Adaptation}} (illustrated in Fig.~\ref{fig:method}), that uses groundtruth correspondences from 3D SfM reconstructions. This section describes how this groundtruth is obtained and exploited.

\subsection{SfM as supervision for feature extraction}
\label{subsec:sfm_supervision}

We run a commonly used SfM method (COLMAP) on real colonoscopy sequences 
and select the subsequences where the reconstruction is successful. 
Our training set contains short sequences ($4$-$7$ sec.), from the complete colonoscopy recordings in EndoMapper dataset,  where COLMAP was able to obtain a 3D reconstruction.

We generate 3D reconstructions for all training sequences with out-of-the-box COLMAP, and we recompute them with features and matches from the official SuperPoint and SuperGlue\footnote{https://github.com/magicleap/SuperGluePretrainedNetwork}. As shown in~\cite{barbed2023tracking}, using two different sources of 3D reconstructions as training data greatly contributes to the performance of our strategy.
We then use both 3D reconstruction sources for the reprojection step.

\begin{figure}[!tb]
    \centering
    \begin{tabular}{@{\;}c@{\;}|@{\;}c@{\;}c@{\;}c@{\;}}
         \includegraphics[width=0.23\linewidth]{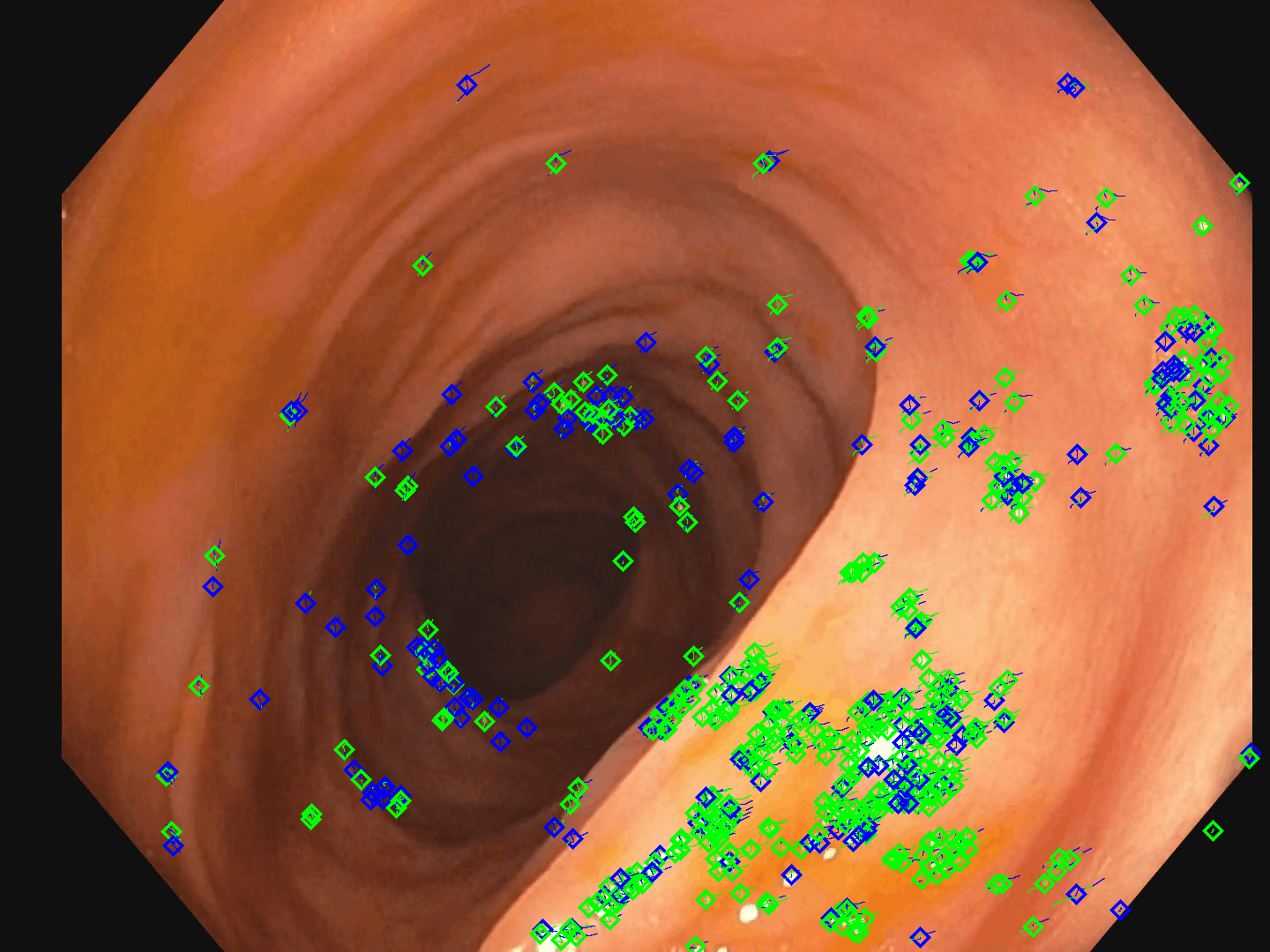} &
         \includegraphics[width=0.23\linewidth]{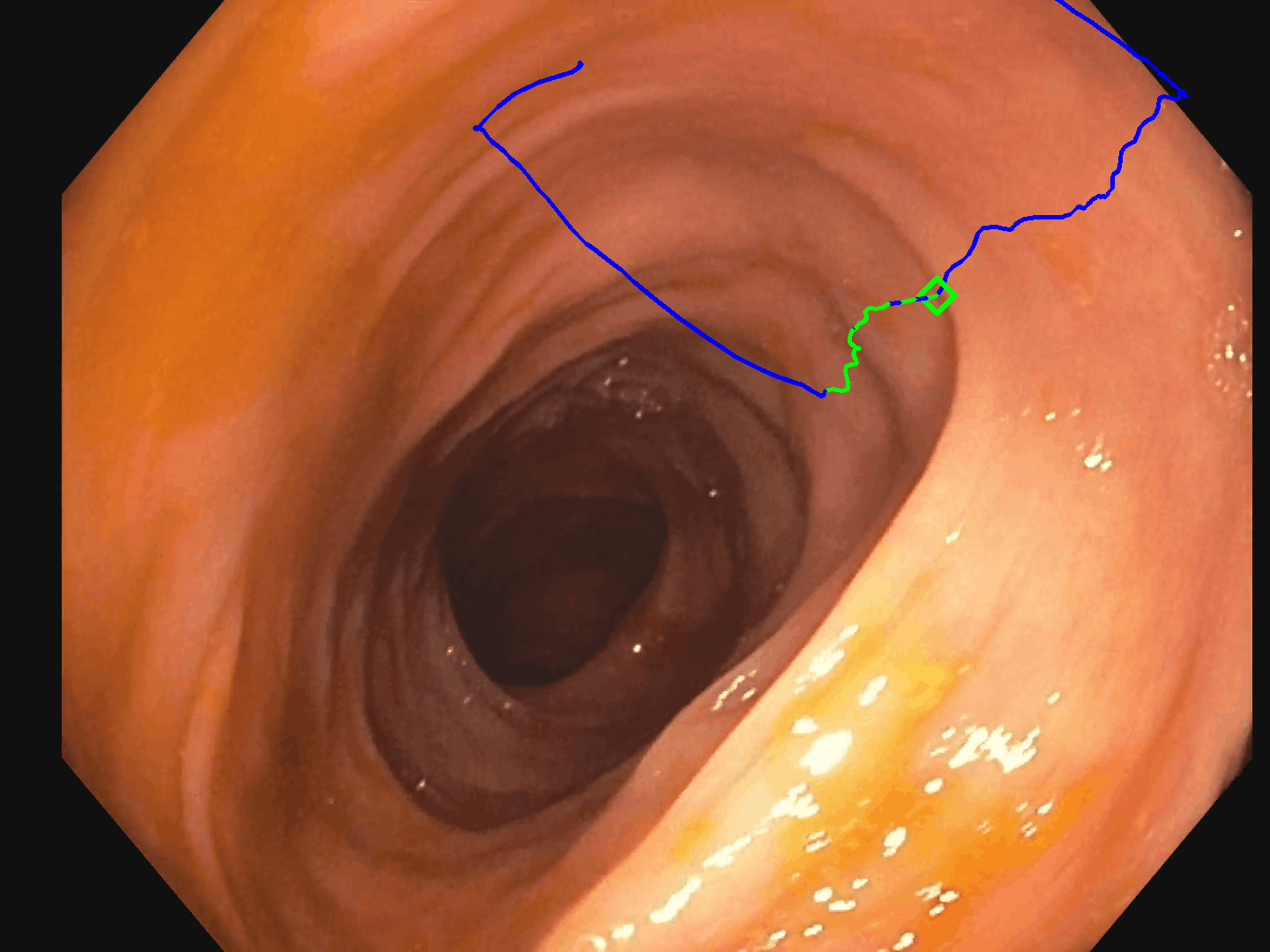} &
         \includegraphics[width=0.23\linewidth]{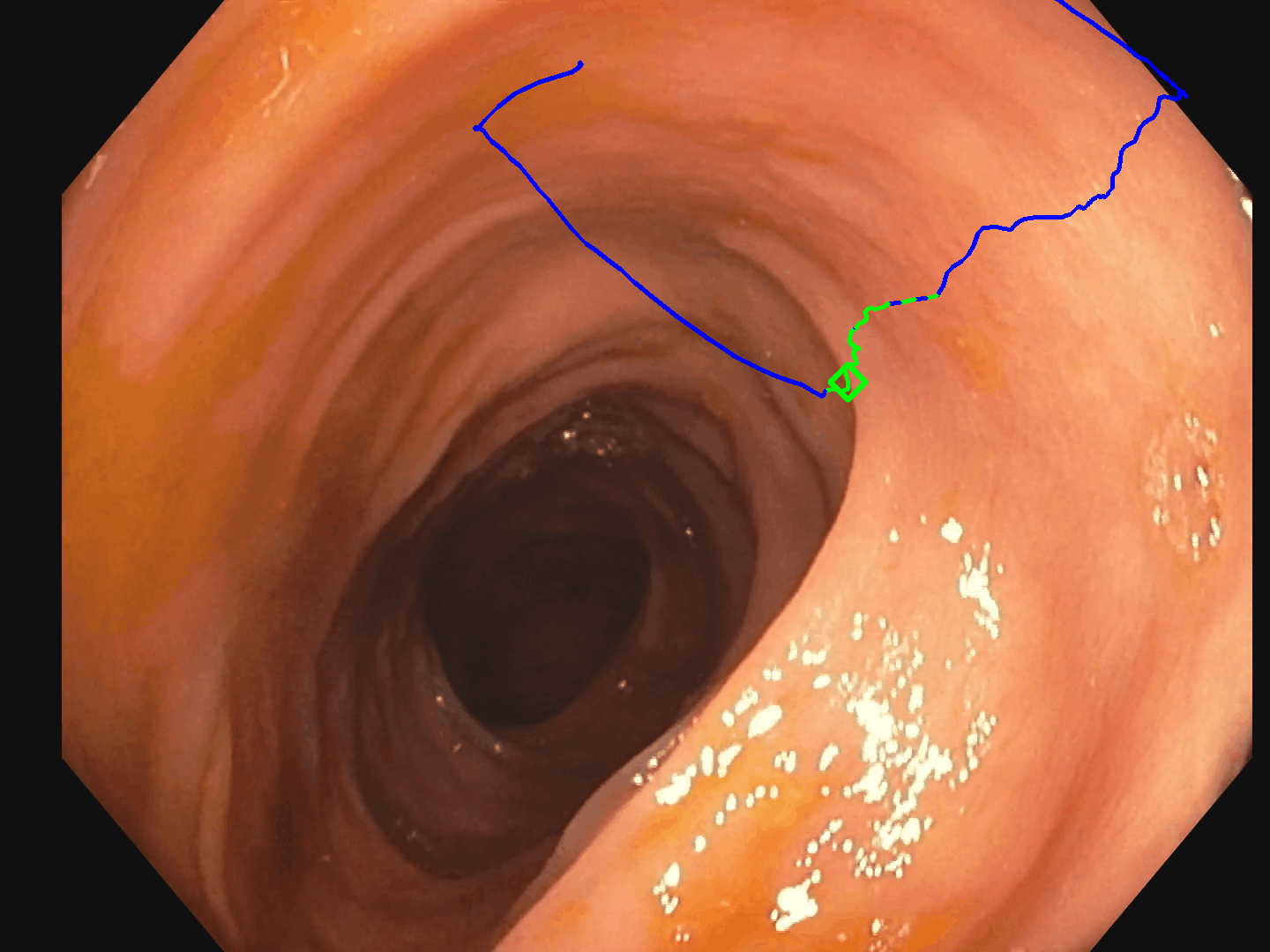} &
         \includegraphics[width=0.23\linewidth]{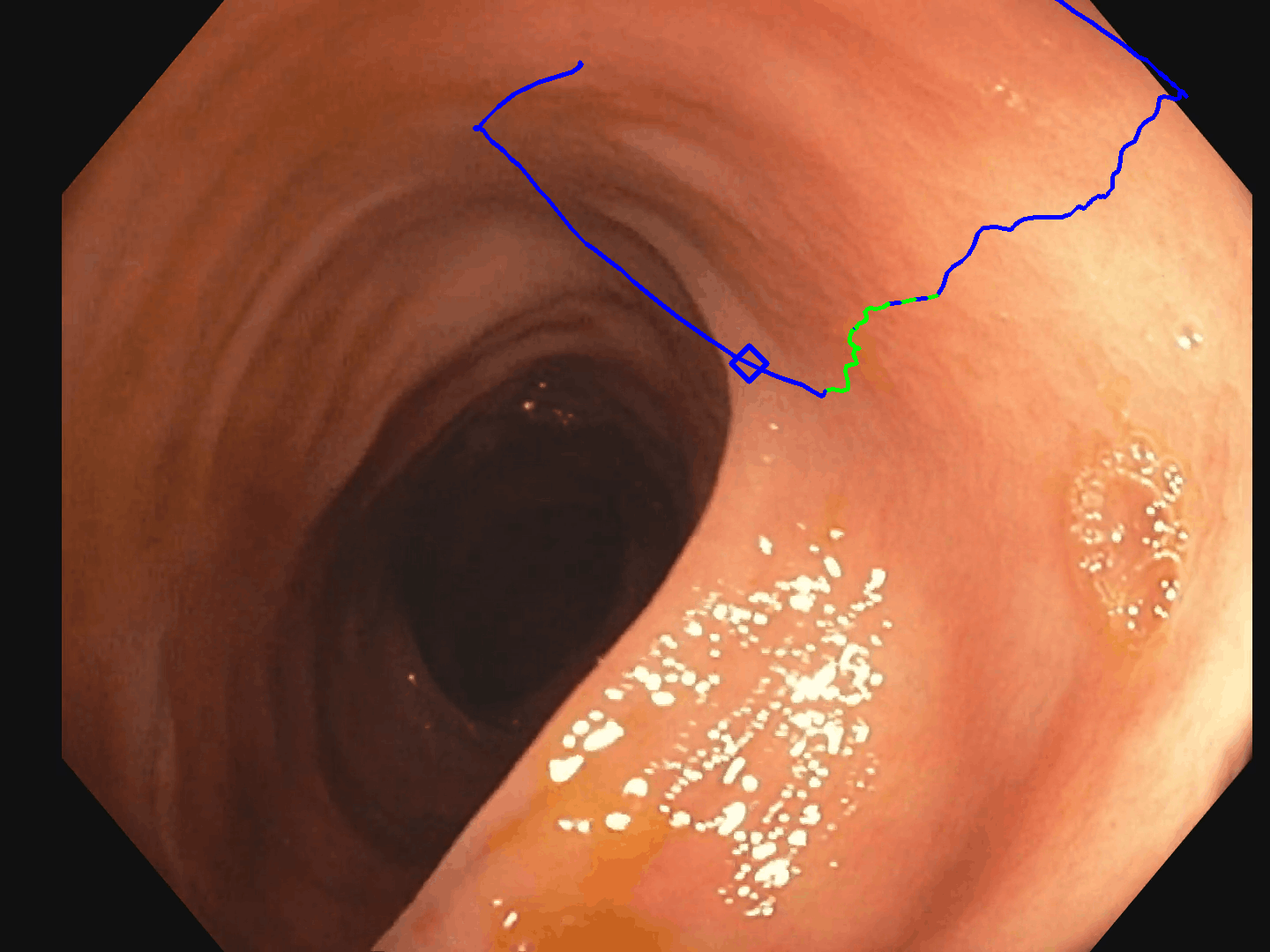} \\
         (a) & (b) & (c) & (d)

    \end{tabular}
    \caption{Supervision points obtained from a COLMAP reconstruction. (a) All 3D points are reprojected into each video frame. Green points were originally detected in this frame, while blue points were not. (b-d) Detail of a complete point track (all positions of one 3D point along the sequence). The \textit{reliable track} for this point is the green segment. The track starts when a point is first detected (b). When the feature is not detected anymore (d), it is depicted in blue from then on and is no longer part of the \textit{reliable track}.}
    \label{fig:supervision_examples}
\end{figure}

We reproject the 3D point cloud from COLMAP into every image. We label the reprojected points in green and blue, as seen in Fig.~\ref{fig:supervision_examples}.
A \textit{reliable track} is an interval of a track bounded by green points. So, the reprojected points selected for training are either green or have preceding and subsequent green points along their tracks. 

The \textit{reliable tracks} are the key of our supervision strategy hypothesis: each point in a \textit{reliable track} corresponds to a different appearance of the same 3D point, i.e., they are correct \textbf{correspondences} between the images. These correspondences supervise the training of our model. 
As it could be expected, there are exponentially more short \textit{reliable tracks} than long ones. This has important design consequences in our approach, such as hyperparameter $N$, used to establish the number of images included in a training batch, as detailed next. 

\subsection{Deep feature extraction for endoscopy}

We use the same network architecture than SuperPoint, with a fully-convolu\-tio\-nal network as encoder backbone and two parallel decoder heads: the detection head that outputs a heatmap of the best pixel locations for feature extraction in the image, and the description head that outputs a 256-long descriptor of floating point numbers for each pixel in the image. 
Our new supervision strategy, named \textit{Tracking Adaptation}, is based on the definition of \textit{reliable tracks} of points and includes a new loss to train SuperPoint that is more suitable for our goal. The original SuperPoint training strategy  uses a synthetic homography to obtain an image $I$ and a warped version $I^{\prime}$. Differently, we use $N$ (hyperparameter of our method) different images from the same sequence that share \textit{reliable tracks} between them to form our training batch, i.e., there is at least one shared \textit{reliable track} between each pair of images in the batch. Our formulation of the loss function, following a similar notation than the original SuperPoint loss $\mathcal{L}_{SP}$, is%
\begin{align}
\begin{split}
   \mathcal{L}_{SPE} (& \mathcal{X}_1,..., \mathcal{X}_N, \mathcal{D}_1,..., \mathcal{D}_N ;Y_1,..., Y_N, 
   \mathcal{T}_{1,2},...,\mathcal{T}_{N-1,N} ) = \\
    &\sum_{n=1}^{N} \mathcal{L}_{p} \left( \mathcal{X}_n, Y_n \right) 
    +
    \lambda \sum_{a=1}^{N-1} \sum_{b=a+1}^{N} \mathcal{L}_{t} \left( \mathcal{D}_a, \mathcal{D}_b, \mathcal{T}_{a,b} \right),
    \label{eq:LSPE}
\end{split}
\end{align}
where $\mathcal{X}_n$ and $\mathcal{D}_n$ are the detection and description heads' outputs from image $I_n$, respectively. $Y_n$ is the supervision for the detection. $\mathcal{T}_{a,b}$ is the correspondence list given the \textit{reliable tracks} shared by every pair of images $I_a$ and $I_b$.

For each image, $I_a$, the corresponding $Y_a$ is the set of 2D image points included in a \textit{reliable track}, whether they are green or blue. 
We maintain SuperPoint's~\cite{detone2018superpoint} detection loss $\mathcal{L}_{p}$. 
Our contribution relies on replacing the description loss $\mathcal{L}_d$ with a new tracking loss that we formulate in similar terms%
\begin{equation}
    \mathcal{L}_{t} \left( \mathcal{D}_a, \mathcal{D}_b, \mathcal{T}_{a,b} \right) = \frac{1}{\left| \mathcal{T}_{a,b}\right| ^2} \sum_{i=1}^{\left| \mathcal{T}_{a,b}\right|} \sum_{j=1}^{\left| \mathcal{T}_{a,b}\right|} l_t \left( \mathbf{d}_{a_{i}}, \mathbf{d}_{b_{j}}, i, j \right),
\end{equation}
\noindent with 
\begin{equation}
   l_t \left( \mathbf{d}_{a_{i}}, \mathbf{d}_{b_{j}}, i, j \right) = \left\{\begin{array}{lr}
        \lambda_t \mbox{max}(0,m_p - \mathbf{d}_{a_{i}}^T \mathbf{d}_{b_{j}}) & \mbox{if } i=j,\\
        \mbox{max}(0,\mathbf{d}_{a_{i}}^T \mathbf{d}_{b_{j}}-m_n) & \mbox{if } i\neq j
        \end{array}\right\},
\end{equation}
where $l_t$ is a triplet loss measuring the distance between positive pairs (weighting parameter $\lambda_t=1$, positive margin $m_p=1$) and the distance between negative pairs (negative margin $m_n=0.2$). Two descriptors from different images, $\mathbf{d}_{a_{i}}$ and $\mathbf{d}_{b_{j}}$, are a positive pair if they belong to the same \textit{reliable track} ($i=j$), and negative pair otherwise ($i\neq j$). 
With this new training loss, our goal is to learn to detect all the points that can be reconstructed, by using their 2D reprojections as supervision for the detection output. Most importantly, our tracking loss $\mathcal{L}_{t}$ helps  the description output to be consistent with the \textit{reliable tracks} used as supervision, i.e.,  all the point appearances from one track have similar descriptors, while being distinct from descriptors from all other tracks.

\section{Experiments}
\label{sec:experiments}
This section compares \mbox{SuperPoint-E} feature quality with SIFT and the original SuperPoint, considering two matching algorithms: Brute Force (BF) and Guided Matching (GM). 

\subsection{Datasets}

Our experiments are run with the \textbf{EndoMapper dataset}~\cite{azagra2023endomapper} of real endoscopy sequences, that contains 96 gastroscopy and colonoscopy procedures recorded at 1440$\times$1080 resolution, with additional meta-data such as camera calibration and voice recording transcriptions. We select a diverse set of sequences for training and testing:
\begin{itemize}
    \item \textit{EM-Train}: 65 subsequences of between 202-342 frames each (4-7~s), a total of 16663 frames, taken from 14 different procedures (sequences 20, 22, 24, 25, 37, 41, 44, 50, 52, 53, 57, 58, 67, 69). They are parts of the sequences where COLMAP is able to make a reconstruction.  For training we use the 11259 frames that COLMAP was able to reconstruct.
    \item \textit{EM-Test}: 7 subsequences of between 105-155 frames each (2-3~s), a total of 838 frames, taken from sequences 1, 2, 14, 16, 17 and two from 95. They are parts of the sequences where COLMAP is able to make a reconstruction. 
    \item \textit{EM-Full}: 5 full sequences of between 3150-7948 frames each (10-33~min), a total of 29371 frames. Sequences 3, 27, 35, 75 and 76. The sequences contain ten times more frames, but we use only one in ten for speed of the experiments and to avoid redundant frames. 
    \item \textit{EM-Gastro}: 5 subsequences of 100 frames each (2~s), a total of 500 frames, from sequence 21 (from the duodenum section), which corresponds to a gastroscopy procedure. 
\end{itemize}

We also use the \textbf{Colonoscopy 3D Video Dataset (C3VD)}~\cite{bobrow2023colonoscopy}. 
This dataset contains recordings made with a real colonoscope in high-fidelity colon models, also recording other values such as depth and camera translation and rotation. 
\textit{C3VD-Test}: 5 subsequences of 100-150 frames each, taken from a diverse set: subsequences cecum\_t2\_a, desc\_t4\_a$^\dagger$, sigmoid\_t3\_a$^\dagger$, trans\_t1\_a, trans\_t2\_b. $^\dagger$ means only one every three frames is used, to have all the sequences with similar number of frames.

\subsection{Methods}
\label{sec:setup_methods}
Following  approaches like~\cite{jin2021image}, we evaluate local features by their impact in downstream tasks. Our experiments combine relevant feature extractors and matching algorithms in the SfM COLMAP pipeline to obtain 3D reconstructions.\\

The \textbf{feature extraction} methods studied are the following:
\begin{itemize}
    \item \textbf{SIFT}~\cite{lowe2004distinctive}. Implementation provided as part of COLMAP. We use the default configuration with GPU acceleration. 
    \item \textbf{SuperPoint} (SP)~\cite{detone2018superpoint}. Official implementation\footnote{https://github.com/magicleap/SuperGluePretrainedNetwork} and weights. Inference parameters: \texttt{\small nms\_radius=}$4$ pixels; \texttt{\small max\_keypoints=}$10\,000$ points; \texttt{\small keypoint\_threshold=}$0.0005$.
    \item \mbox{ \bf SuperPoint-E} (SP-E). Our model is a modified implementation of SuperPoint~\cite{jau2020deep}\footnote{https://github.com/eric-yyjau/pytorch-superpoint}, trained on \textit{EM-Train}. An ablation study on the training strategy proposed can be found in~\cite{barbed2023tracking}. 
\end{itemize}
\noindent \mbox{SuperPoint-E} \textbf{training} configuration. The video frames are center-cropped to square shape and resized to 256$\times$256 resolution, converted to grayscale and normalized dividing its values by $255$. We use image augmentation techniques such as: random brightness of $\pm50$ intensity value, random contrast variation changing the intensity value for $127 + \alpha(v-127)$ where $v$ is the pixel intensity value and $\alpha \sim U(0.5,1.5)$, additive speckle noise with a probability $p \sim U(0,0.0035)$ and additive Gaussian noise with $\sigma \sim U(0,10)$, adding random elliptical shades, and motion blur with kernel size of $3$~pixels. 
A Gaussian blur is applied on top of the point reprojections in $Y_i$ to smooth loss values with $\sigma=0.2$~pixels. On the detection head we use a detection threshold of $0.015$ and Non-Maximum Suppression of radius $4$~pixels. The parameters for the loss are $\lambda=1$, $\lambda_t=1$, $m_p=1$ and $m_n=0.2$. The model is trained with batch size $N=4$~images for 400\,000 training batches and a learning rate of $1\mathrm{e}{-5}$.

\noindent \mbox{SuperPoint-E} \textbf{testing} configuration: The same parameters as for SuperPoint are used for a fair comparison, with \texttt{\small max\_keypoints=}$10\,000$~points, \texttt{\small keypoint\_threshold=}$0.0005$ and \texttt{\small nms\_radius=}$4$~pixels.\\


\textbf{Feature matching} methods. To study the feature extraction quality without getting influence from very sophisticated matching networks, tailored often to a specific detector, we consider two generic matching strategies run as part of COLMAP: Brute Force and Guided Matching. 

\noindent Exhaustive matching, or \textbf{Brute Force} (BF), 
computes the cosine similarity between each feature descriptor from an image and all the descriptors from another image. The descriptor of the other image that has the highest similarity is chosen as the match. By default, the option \texttt{\small cross\_check=}$true$ makes this matching bi-directional, i.e., it is counted as correct if and only if the match appears simultaneously from image $I_a$ to image $I_b$ and vice versa. The default parameters are optimal for SIFT. For  SuperPoint models we set \texttt{\small max\_ratio=}$1$ and \texttt{\small max\_distance=}$1$~rad.


\noindent The \textbf{Guided Matching} (GM) strategy builds upon the BF method. After getting the bi-directional matches between two images ($I_a$ and $I_b$), this strategy estimates the most probable Fundamental Matrix ($F_{ab}$) that is consistent with the displacement of the majority of the matches using the RANSAC algorithm. $F_{ab}$ is then used to \textit{guide} a second matching round, where each point $p_a$ from $I_a$ is transformed to project it onto $I_b$'s image space: $p^{\prime}_a = F_{ab} p_a$. Finally, if there is a point $p_b$ in $I_b$ that is closer to $p^{\prime}_a$ than \texttt{\small max\_error=}$4$ pixels, it is considered a correct match. The rest of  the parameters are the same as with BF. This process enforces a global consistency model, which typically produces better sets of matches between images than the previous method.
\\

\textbf{Triangulation \& Bundle Adjustment} is the final step where COLMAP processes the sets of matches between all the images and builds the largest possible submaps (reconstructions, models). All parameters are set as default except \texttt{\small init\_min\_tri\_angle=}$8$, \texttt{\small ba\_refine\_focal\_length=}$0$ and \texttt{\small ba\_refine\_extra\_params=}$0$.

\subsection{Training supervision from COLMAP}

The COLMAP configuration to generate 3D reconstructions for all training sequences uses these blocks and parameters: 
\textit{feature\_extractor} with \texttt{\small OPENCV\_FISHEYE} camera model and calibration from endoscope Nº1 in~\cite{azagra2023endomapper}; \textit{exhaustive\_matcher} with \texttt{\small guided\_matching=}$1$; and \textit{mapper} with \texttt{\small init\_min\_tri\_angle=}$8$~deg and  \texttt{\small min\_model\_size=}$50$~images. 
As shown in the experiments in~\cite{barbed2023tracking}, using two different sources of 3D reconstructions as training data greatly contributes to the performance of our strategy. 
Therefore, we additionally compute the 3D reconstruction for these same sequences with a modified COLMAP pipeline that uses 
the official SuperPoint and SuperGlue\footnote{https://github.com/magicleap/SuperGluePretrainedNetwork} implementation, with the available \textit{indoor} set of weights. 
We set default parameter values except \texttt{\small keypoint\_threshold=}$0.015$ and \texttt{\small nms\_radius=}$1$~pixel. 
The SuperGlue resulting matches are directly introduced into COLMAP, and only the \textit{mapper} module is run, with the same configuration as before. 

\subsection{Experiments}

\subsubsection{Discriminative and repeatable features} 
\label{subsec:exp_quality}
This experiment extracts  features on the short (2-3 s) \textit{EM-Test} sequences with all the methods considered, and feeds them to COLMAP to obtain the corresponding scene 3D reconstruction. These reconstructions are represented by the computed camera locations at each frame, and the 3D point cloud with all reconstructed points. 
To evaluate desirable aspects of the features for 3D reconstruction we measure:
\begin{itemize}
    \item \textbf{Precision}:  average percentage of detected features in an image that are successfully triangulated and part of the final 3D point cloud.
    \item \textbf{Reconstructed images}:  percentage of images included in the final reconstruction. 
    \item \textbf{3D points}:  number of 3D points in the final reconstruction.
    \item \textbf{Track length}: average number of images that each point is triangulated from, i.e., it was identified as the same feature in all those images. 
    \item Mean Absolute Error (\textbf{MAE}): mean reprojection error of all reconstructed points. The mean of the ten thousand points with lowest reprojection error is \textbf{MAE 10K}.
    \item \textbf{Spread}: measure of the spread of the reconstructed features over the images. We divide the images into a 16x16 grid and find the percentage of the resulting cells that have at least one reconstructed feature inside.
    \item \textbf{Specular}: percentage of reconstructed features localized on top of a specular reflection. A pixel is considered part of a specular reflection if intensity$\geq180$, with $0\leq\mbox{intensity}\leq255$.
\end{itemize}
\begin{table}[!tb]
    \centering
    \footnotesize
    \caption{Reconstruction quality metrics (average over the 7 sequences in \textit{EM-Test})}
    \begin{tabular}{@{}l@{}|@{\hspace{1mm}}c@{\hspace{1mm}}|@{\hspace{1mm}}c@{\hspace{1mm}}|@{\hspace{1mm}}c@{\hspace{1mm}}|@{\hspace{1mm}}c@{\hspace{1mm}}|@{\hspace{1mm}}c@{\hspace{1mm}}} 
        \hline
         & SIFT+GM & SP+BF & SP-E+BF & SP+GM & SP-E+GM \\ 
         & COLMAP &  & (Ours) &  & (Ours) \\ 
        \hline
        \textbf{Precision} (\%points) $\uparrow$ & $46.1$\% & $40.6$\% & $60.5$\% & $57.7$\% & $\textbf{63.2}$\% \\ 
        \textbf{Reconstr.} (\%imgs) $\uparrow$ & $87.1$\% & $96.0$\% & $99.5$\% & $\textbf{100}$\% & $\textbf{100}$\% \\ 
        \textbf{3D points}* $\uparrow$ & $10$K & $22$K & $76$K & $49$K & $\textbf{77}$K \\ 
        \textbf{Track-length}(imgs) $\uparrow$ & $9.12$ & $7.05$ & $10.78$ & $5.02$ & $\textbf{11.28}$ \\ 
        \textbf{MAE} (pixels) $\downarrow$ & $\textbf{1.31}$ & $1.55$ & $1.78$ & $1.44$ & $1.79$ \\ 
        \textbf{MAE 10K} (pixels) $\downarrow$ & $1.20$ & $1.05$ & $0.69$ & $\textbf{0.44}$ & $0.70$ \\ 
        \textbf{Spread} (\%cells) $\uparrow$ & $43.9$\% & $72.3$\% & $85.2$\% & $\textbf{91.7}$\% & $86.3$\% \\ 
        \textbf{Specular} (\%points) $\downarrow$ & $28.6$\% & $15.6$\% & $\textbf{6.2}$\% & $11.3$\% & $6.7$\% \\ 
        \hline
        \multicolumn{6}{l}{* \textbf{3D points} metric corresponds to thousands of points reconstructed.}\\
    \end{tabular}
    \label{tab:exp_quality}
\end{table}

\begin{figure}[!tb]
    \centering
    \footnotesize
    \begin{tabular}{@{}c@{\,}c@{\,}c@{\,}c@{}}
         Original & SIFT+GM & SP+GM & SP-E+GM (Ours) \\       
         \includegraphics[width=0.24\linewidth]{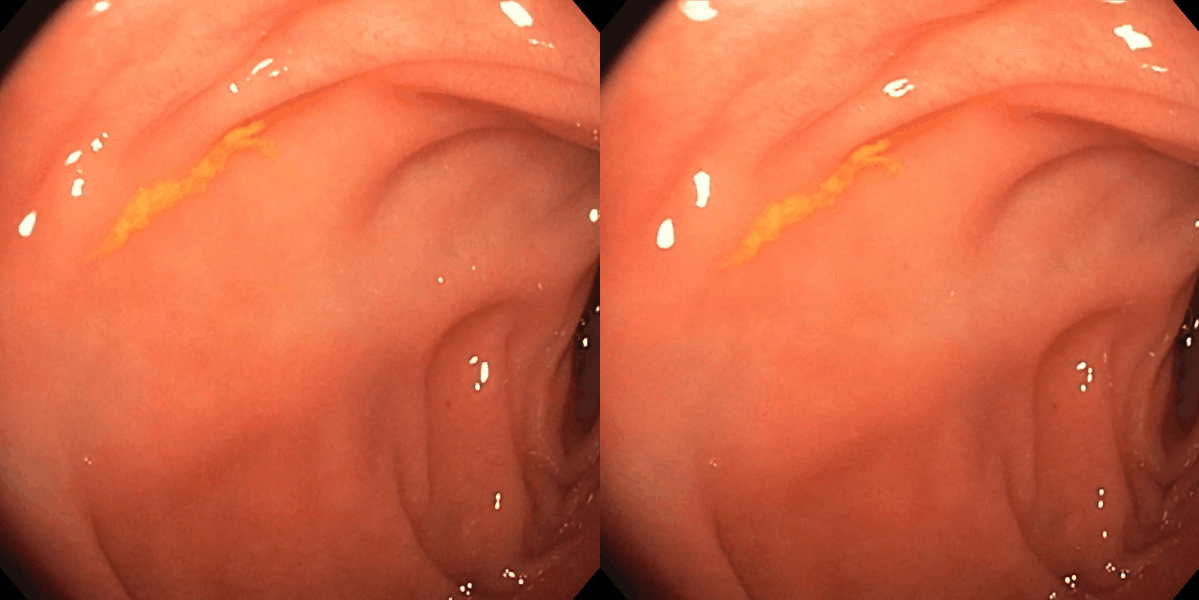} &
         \includegraphics[width=0.24\linewidth]{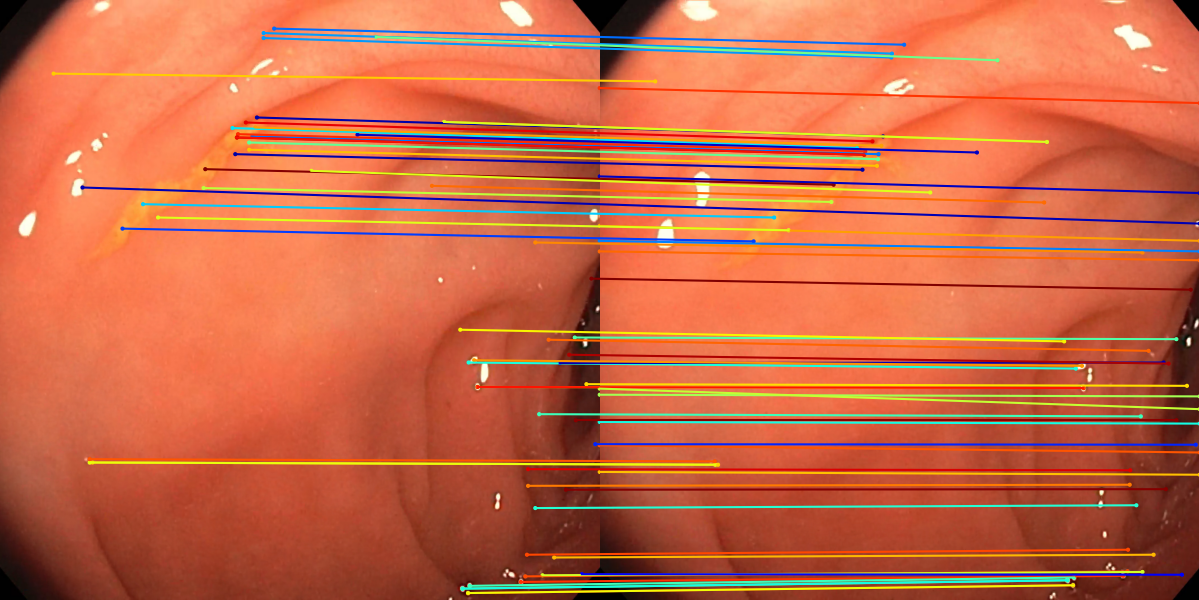} &
         \includegraphics[width=0.24\linewidth]{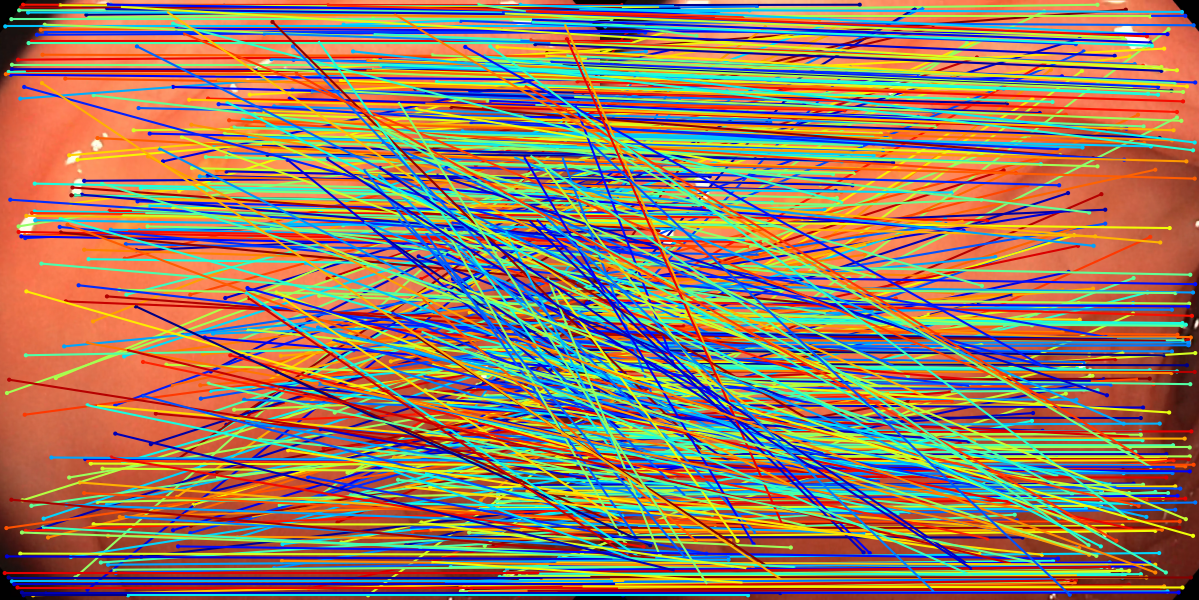} &
         \includegraphics[width=0.24\linewidth]{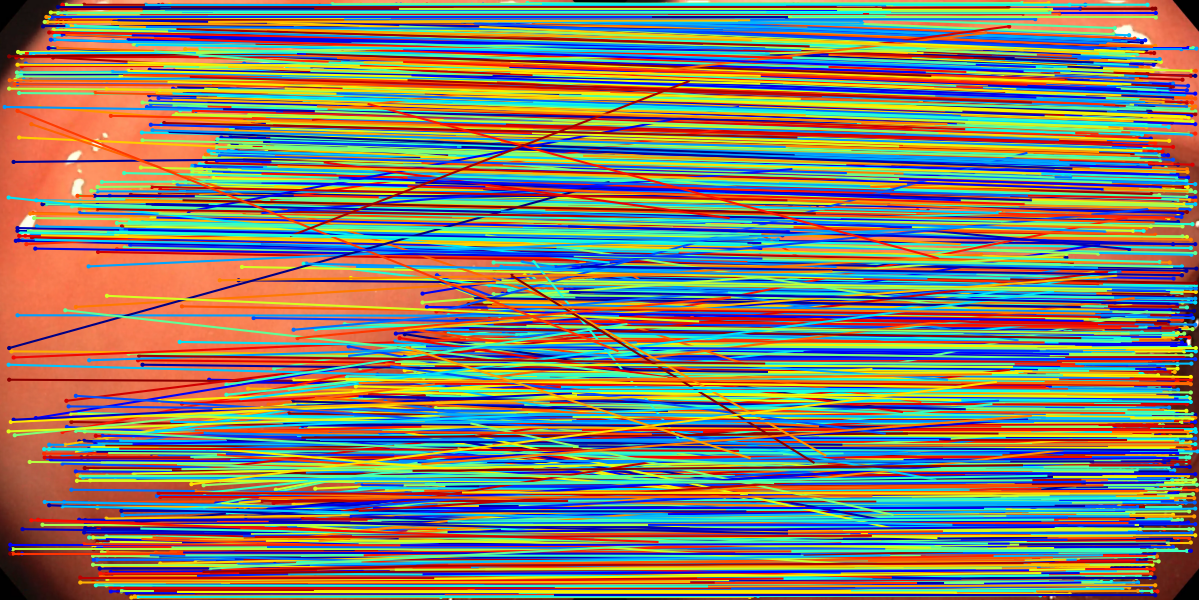} \\
         \includegraphics[width=0.24\linewidth]{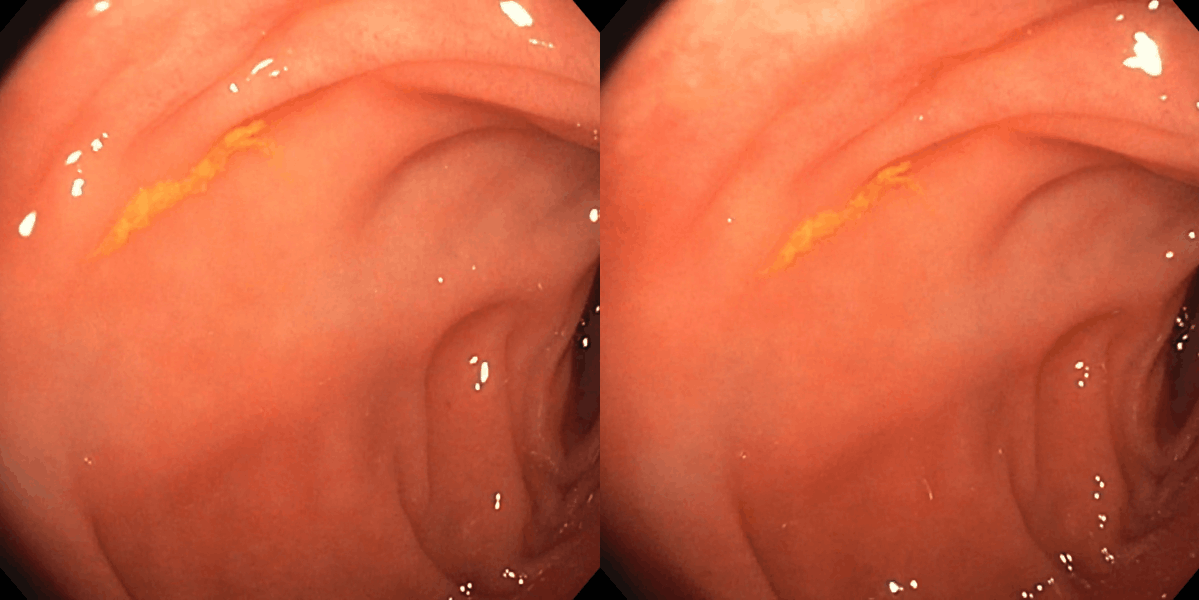} &
         \includegraphics[width=0.24\linewidth]{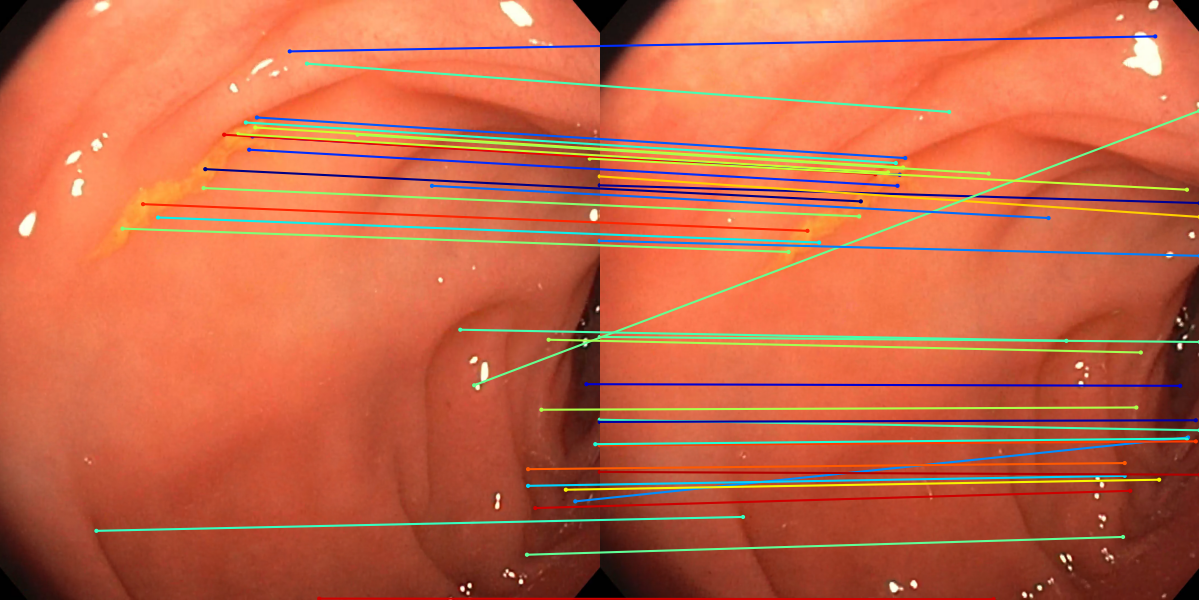} &
         \includegraphics[width=0.24\linewidth]{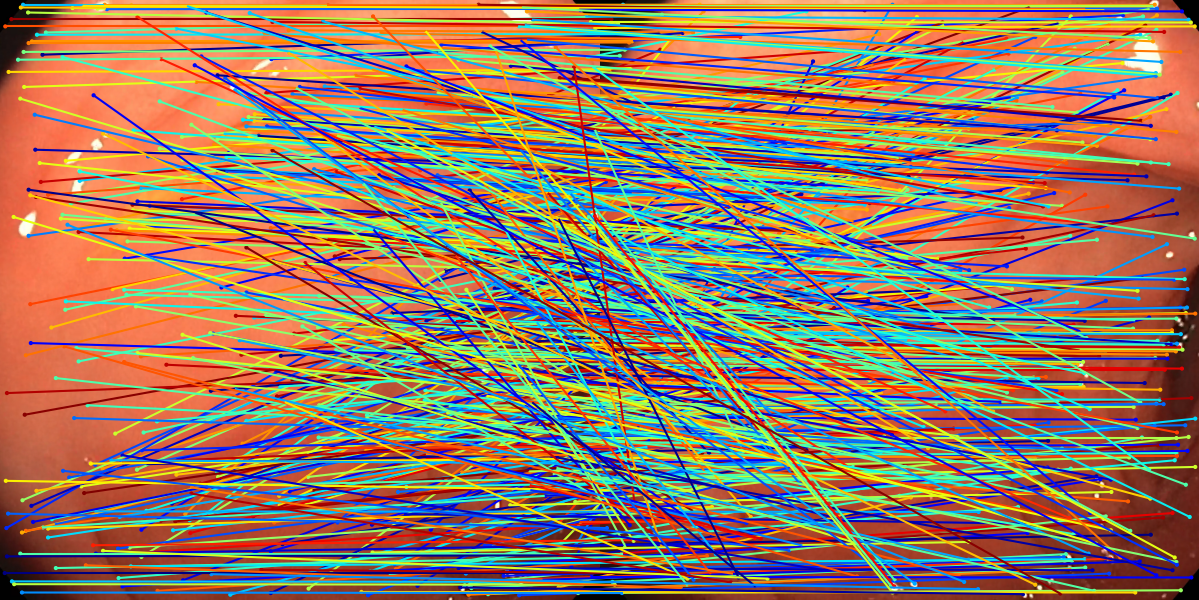} &
         \includegraphics[width=0.24\linewidth]{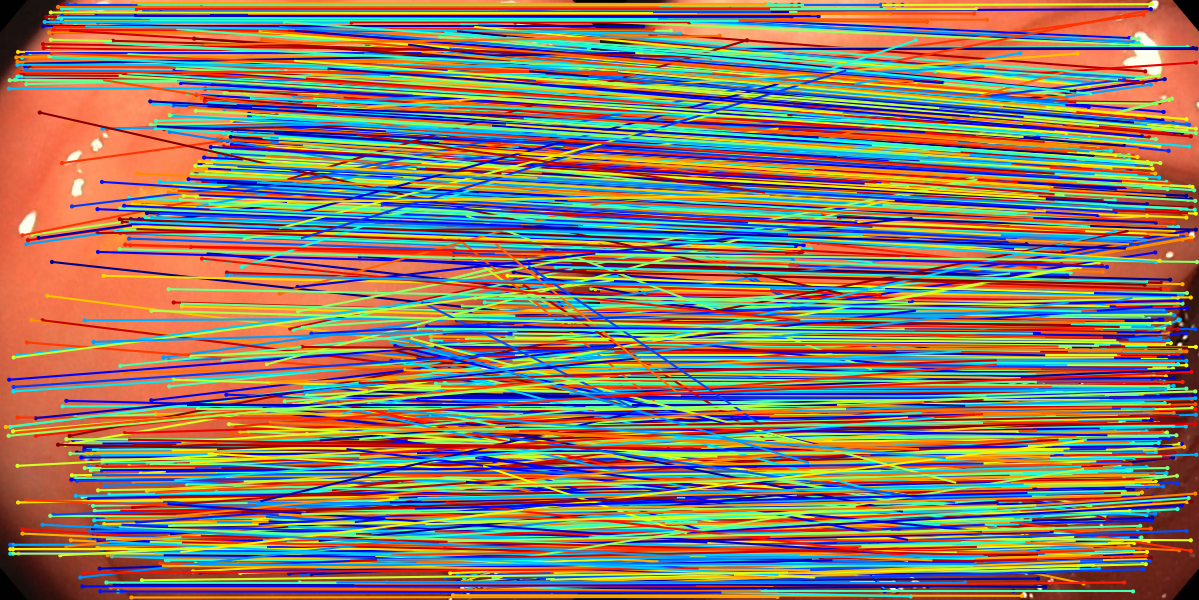} \\
         \includegraphics[width=0.24\linewidth]{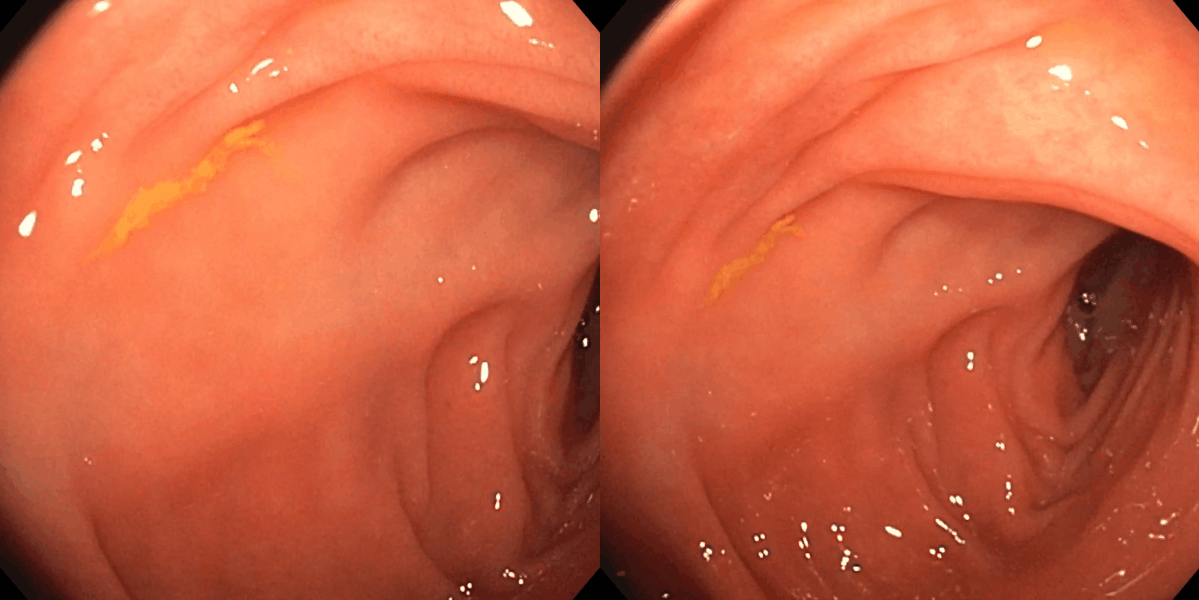} &
         \includegraphics[width=0.24\linewidth]{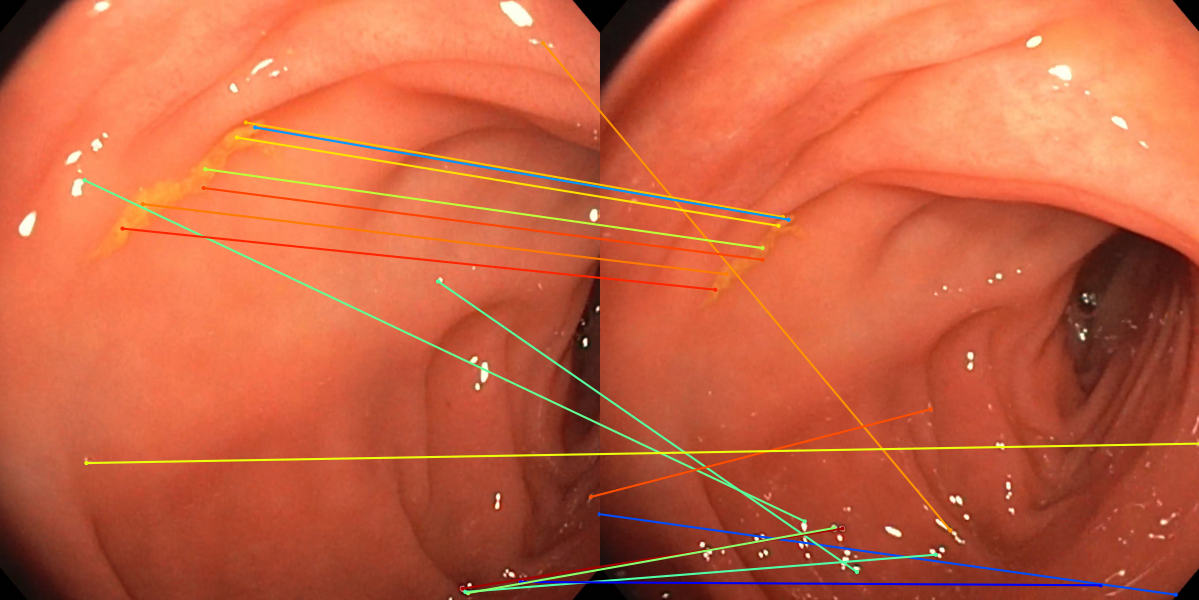} &
         \includegraphics[width=0.24\linewidth]{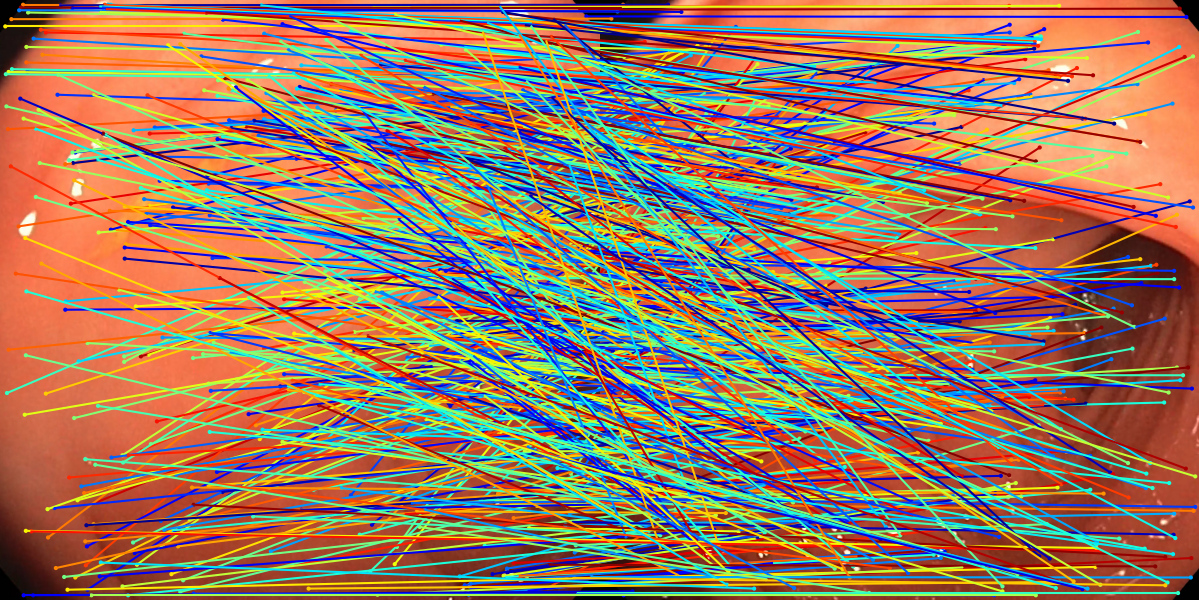} &
         \includegraphics[width=0.24\linewidth]{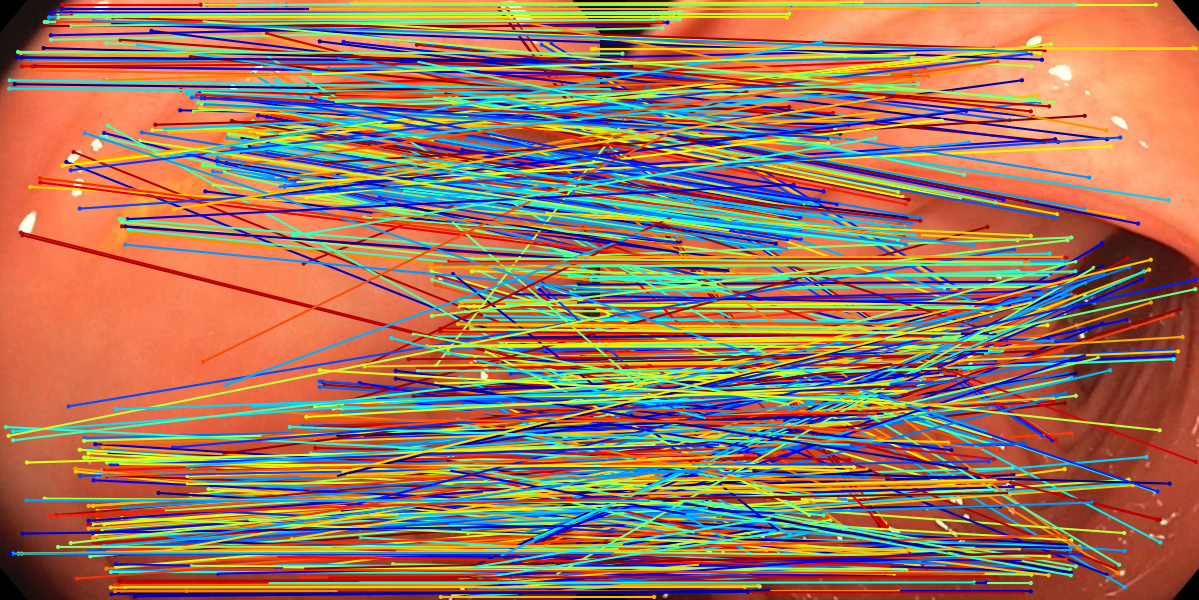} \\
         \hline \noalign{\global\arrayrulewidth=1mm}\arrayrulecolor{white}\hline \noalign{\global\arrayrulewidth=0.4pt}
         \includegraphics[width=0.24\linewidth]{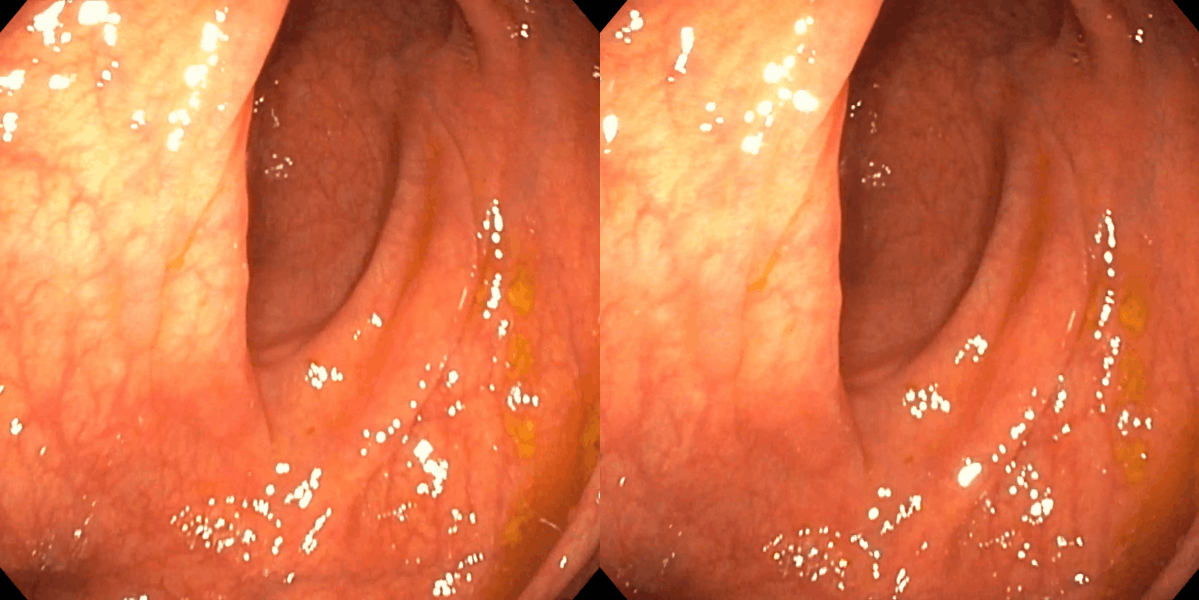} &
         \includegraphics[width=0.24\linewidth]{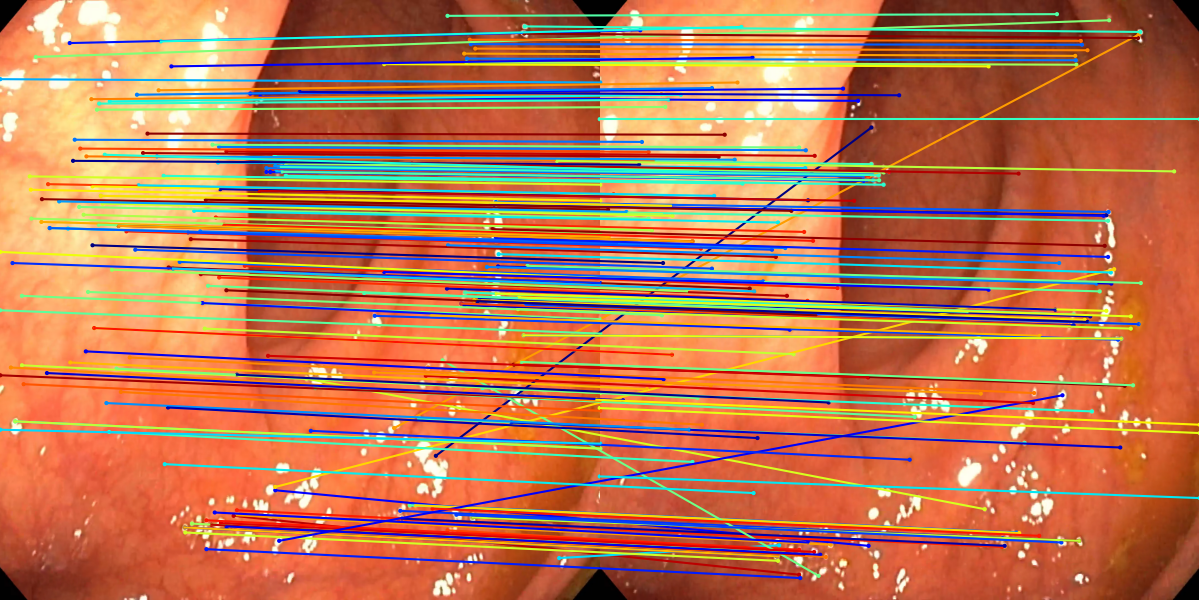} &
         \includegraphics[width=0.24\linewidth]{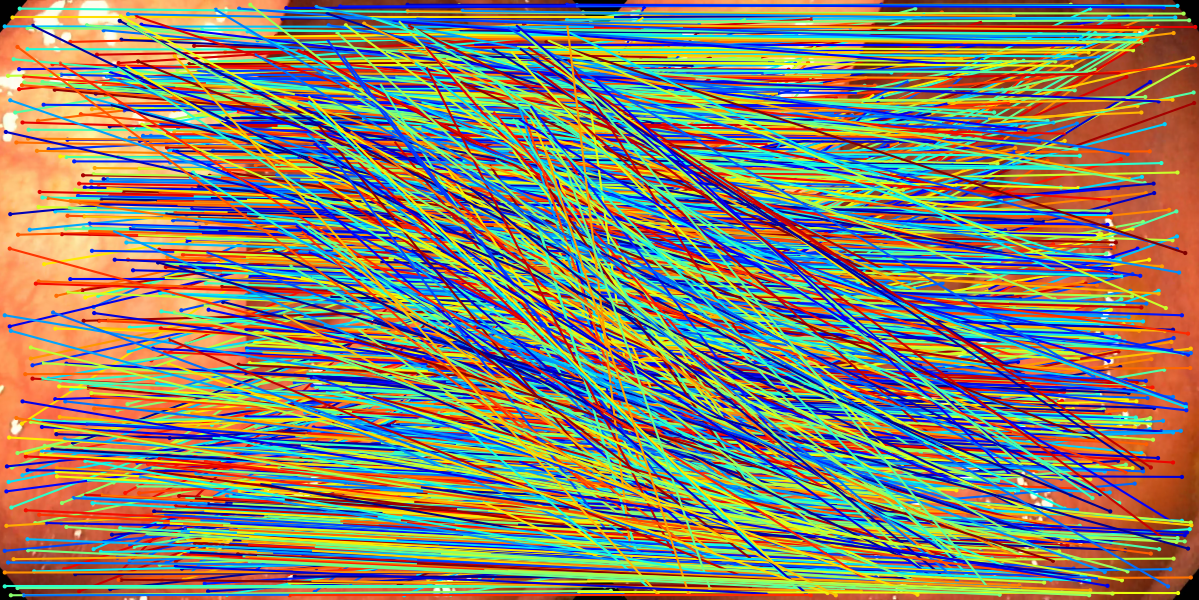} &
         \includegraphics[width=0.24\linewidth]{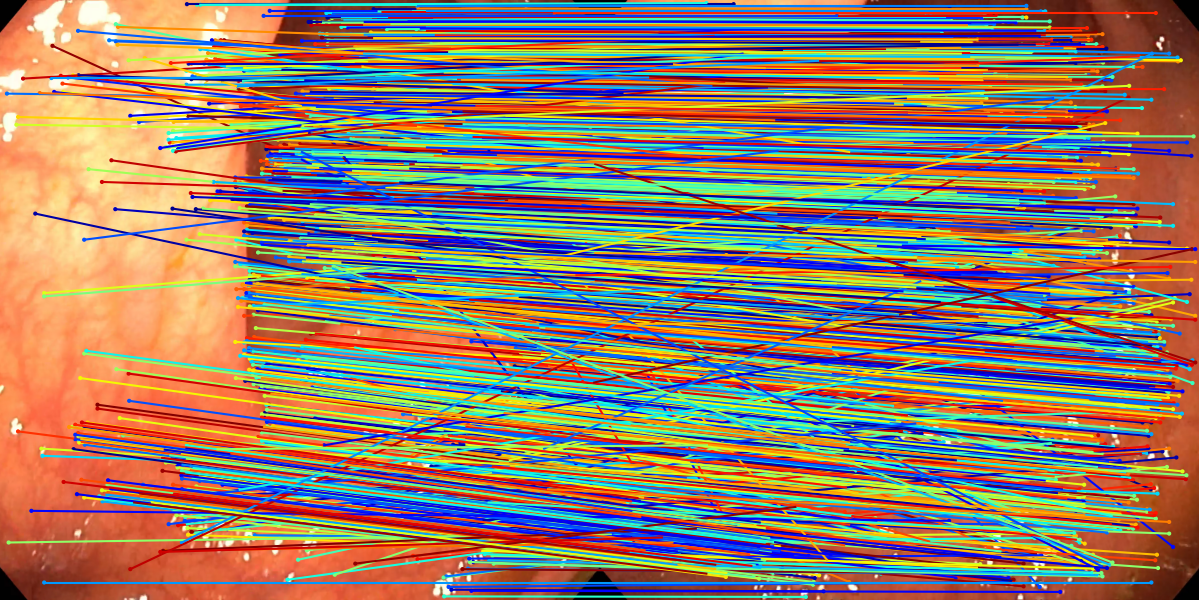} \\
         \includegraphics[width=0.24\linewidth]{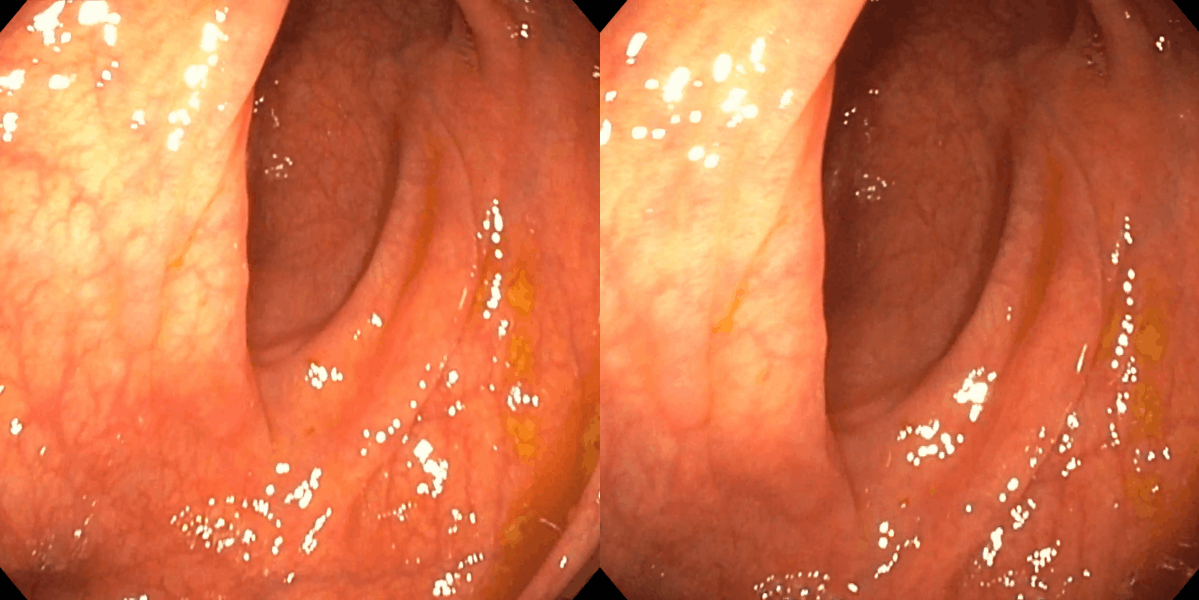} &
         \includegraphics[width=0.24\linewidth]{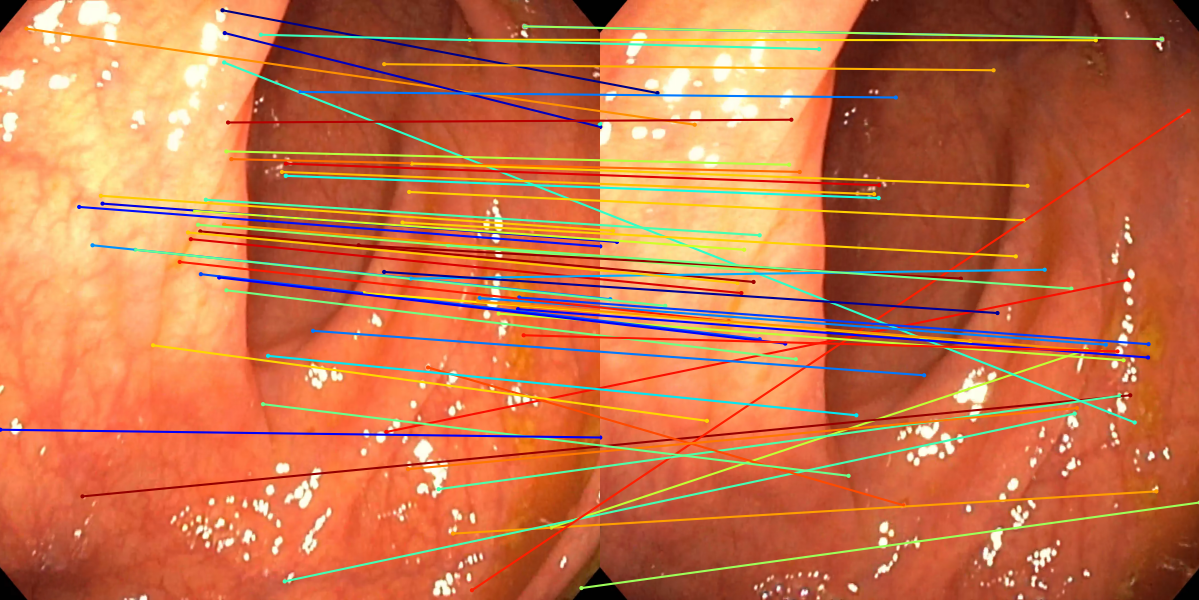} &
         \includegraphics[width=0.24\linewidth]{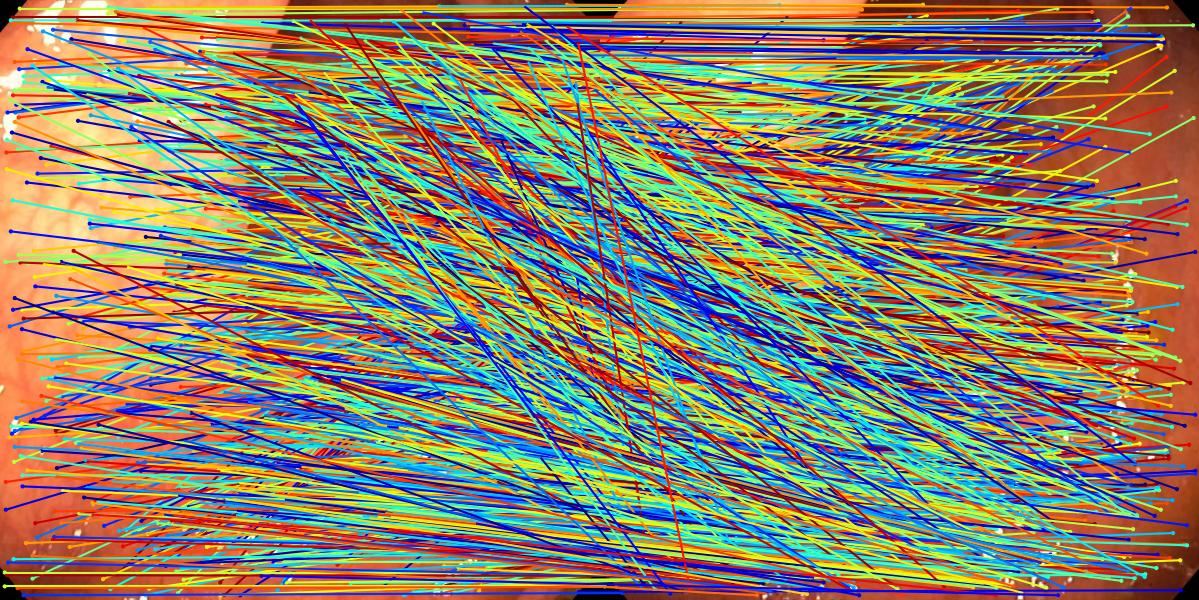} &
         \includegraphics[width=0.24\linewidth]{images/matching/SPE_GM_Seq_095_018_020_Seq_095_018_030.png} \\
         \includegraphics[width=0.24\linewidth]{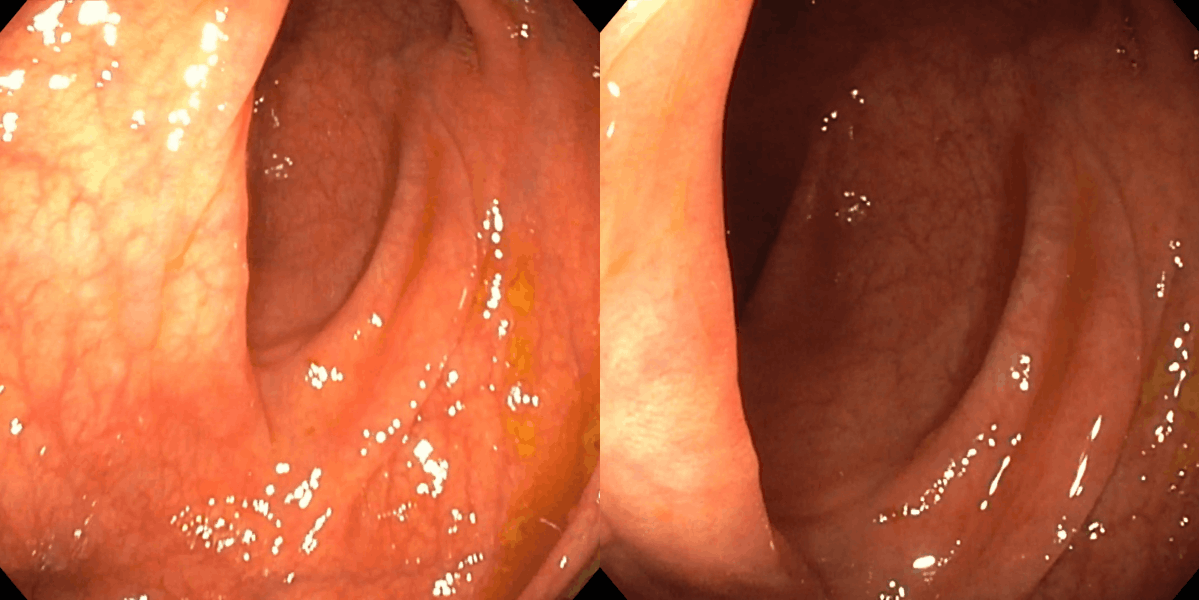} &
         \includegraphics[width=0.24\linewidth]{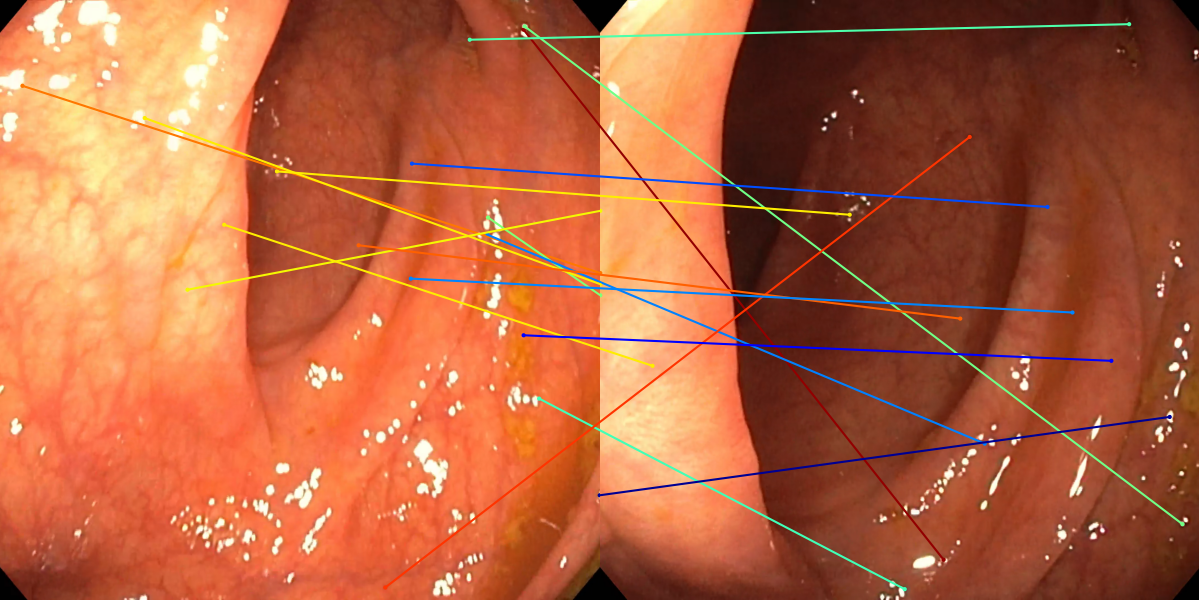} &
         \includegraphics[width=0.24\linewidth]{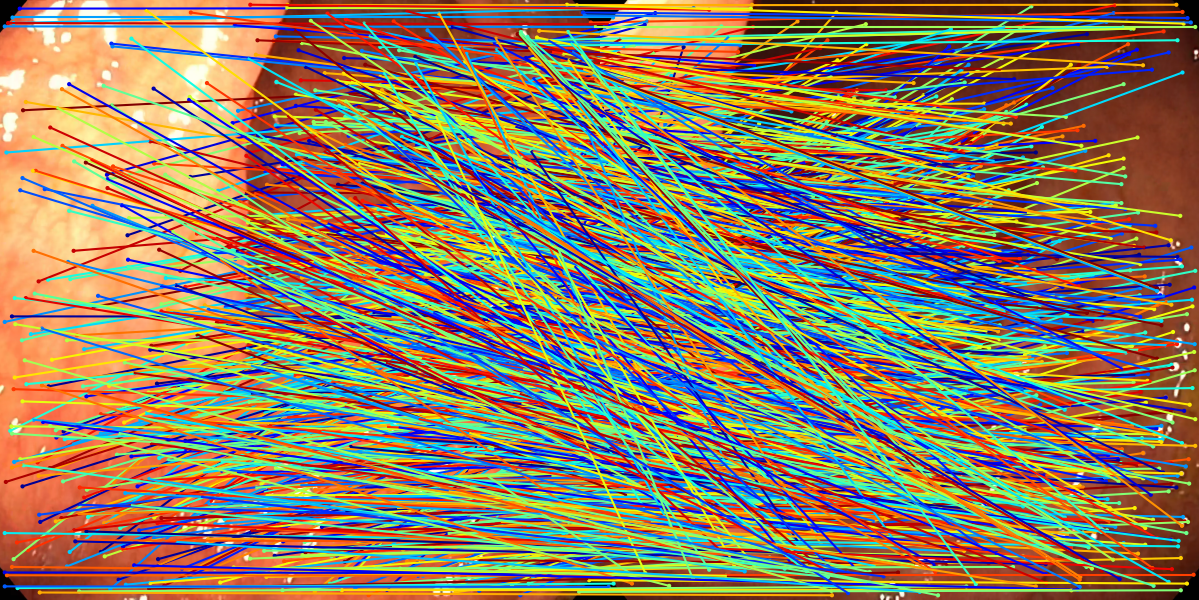} &
         \includegraphics[width=0.24\linewidth]{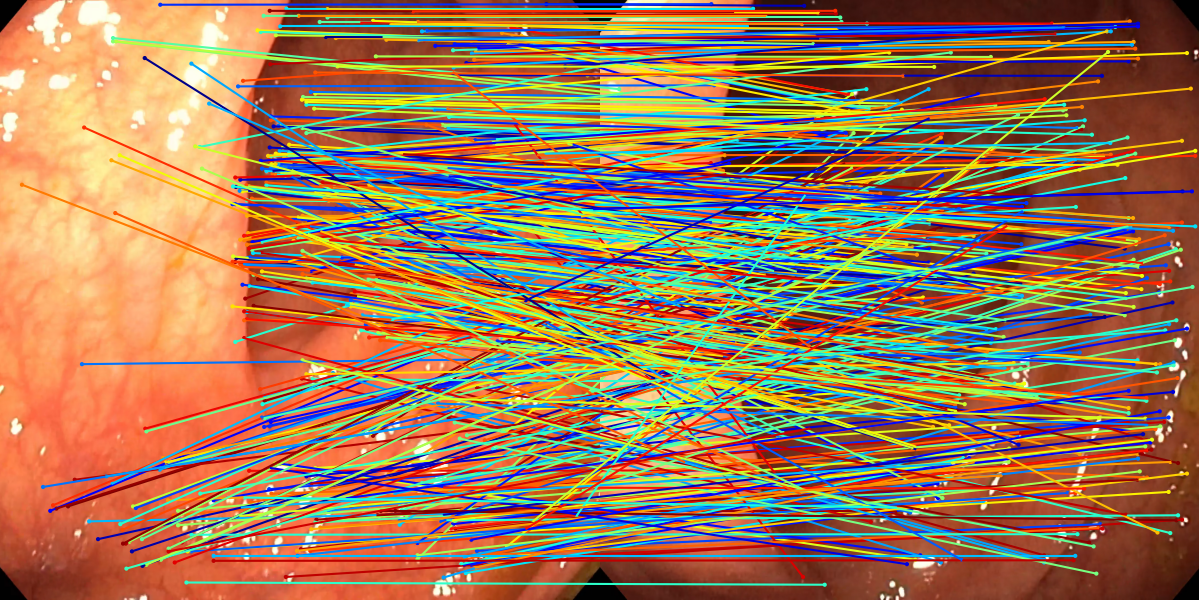} \\
    \end{tabular}
    \caption{Matching example of pairs of frames, 0.1, 0.2 and 0.5 seconds apart within two sequences in \textit{EM-Test}. }
    \label{fig:matching_examples}
\end{figure}

\noindent
The experiment results, summarized in Table~\ref{tab:exp_quality}, show that the proposed \mbox{SuperPoint-E} (SP-E) using GM is the best performing combination of detector and matching algorithm (see some matching examples in Fig.~\ref{fig:matching_examples}). 
We obtain relevant improvements of SuperPoint using our training strategy. \cite{barbed2023tracking} demonstrate how the original SuperPoint (SP) fine-tuned in the target image domain is not sufficient to reach SP-E's performance. SP-E outperforms SP in \textit{Precision}, \textit{3D points}, \textit{Track length} with comparable \textit{Spread} and much lower \textit{Specular}.
In fact, our model presents unique robustness to specular reflections, obtaining the lowest \textit{Specular} percentage ($6.2\%$, $6.7\%$) of all methods by a significant margin. 
Our SP-E model extracts descriptors discriminative enough for a simple BF matching strategy to work as good as the more refined but costly GM method. 
Interestingly, SP-E is not as dependant on it, obtaining similar metrics with either matching method: \textit{Precision} ($60.5\%$, $63.2\%$) and \textit{3D points} ($76$K, $77$K) to name a few. 
Current deep learning-based matching approaches \cite{sarlin2020superglue,lindenberger2023lightglue} improve SuperPoint performance but they require accurate and dense 3D reconstruction groundtruth for training, which currently does not exist in real endoscopy. 
\begin{figure}[!tb]
    \footnotesize
    \centering
    \begin{tabular}{@{}c@{}}
         Subsequence Seq\_001\_1\\ 
         \begin{minipage}{0.42\linewidth}
         \begin{tabular}{@{}c@{}}
         \includegraphics[width=\linewidth]{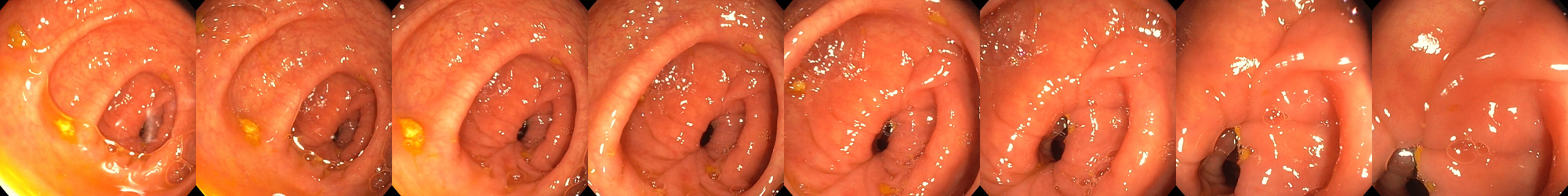} \\
         \includegraphics[width=\linewidth]{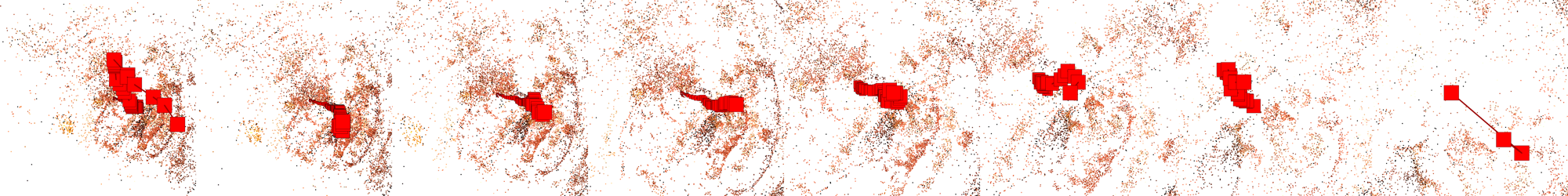} \\
         \includegraphics[width=\linewidth]{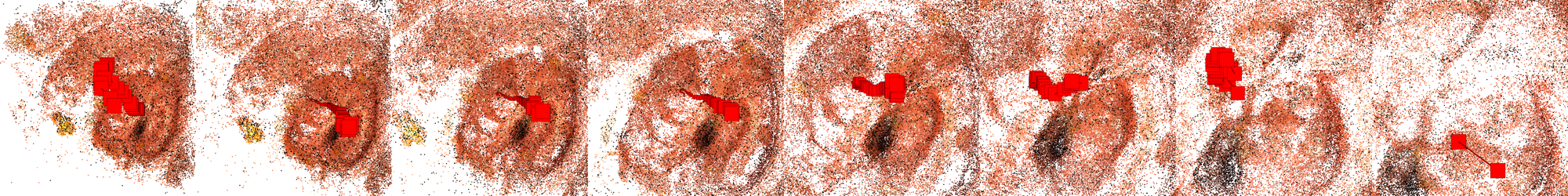} \\
         \end{tabular}
         \end{minipage}
         \begin{minipage}{0.42\linewidth}
         \begin{tabular}{@{}cc@{}}
         \includegraphics[width=0.40\linewidth,trim=10cm 2cm 20cm 8cm,clip]{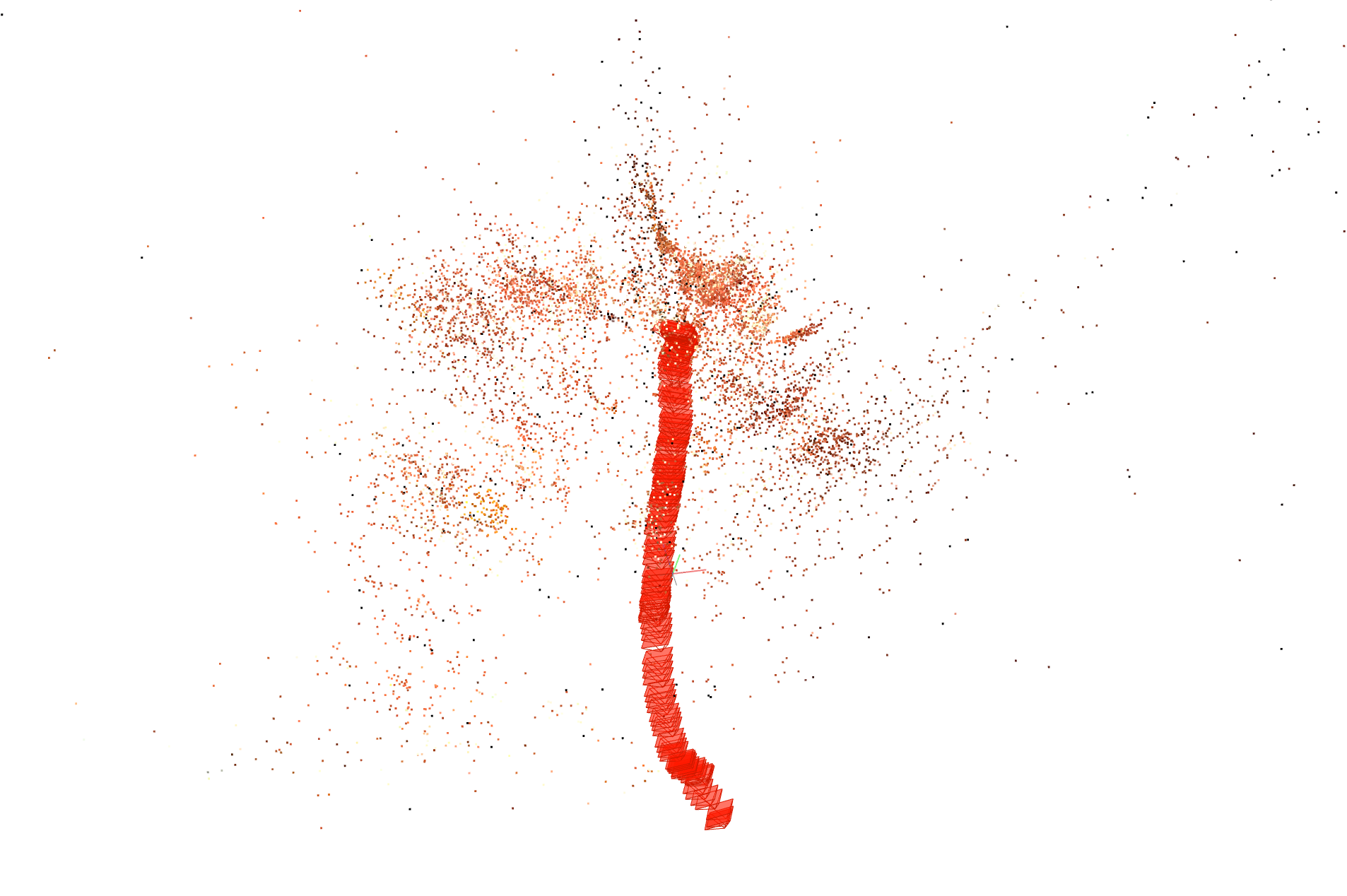} &
         \includegraphics[width=0.40\linewidth,trim=10cm 2cm 20cm 8cm,clip]{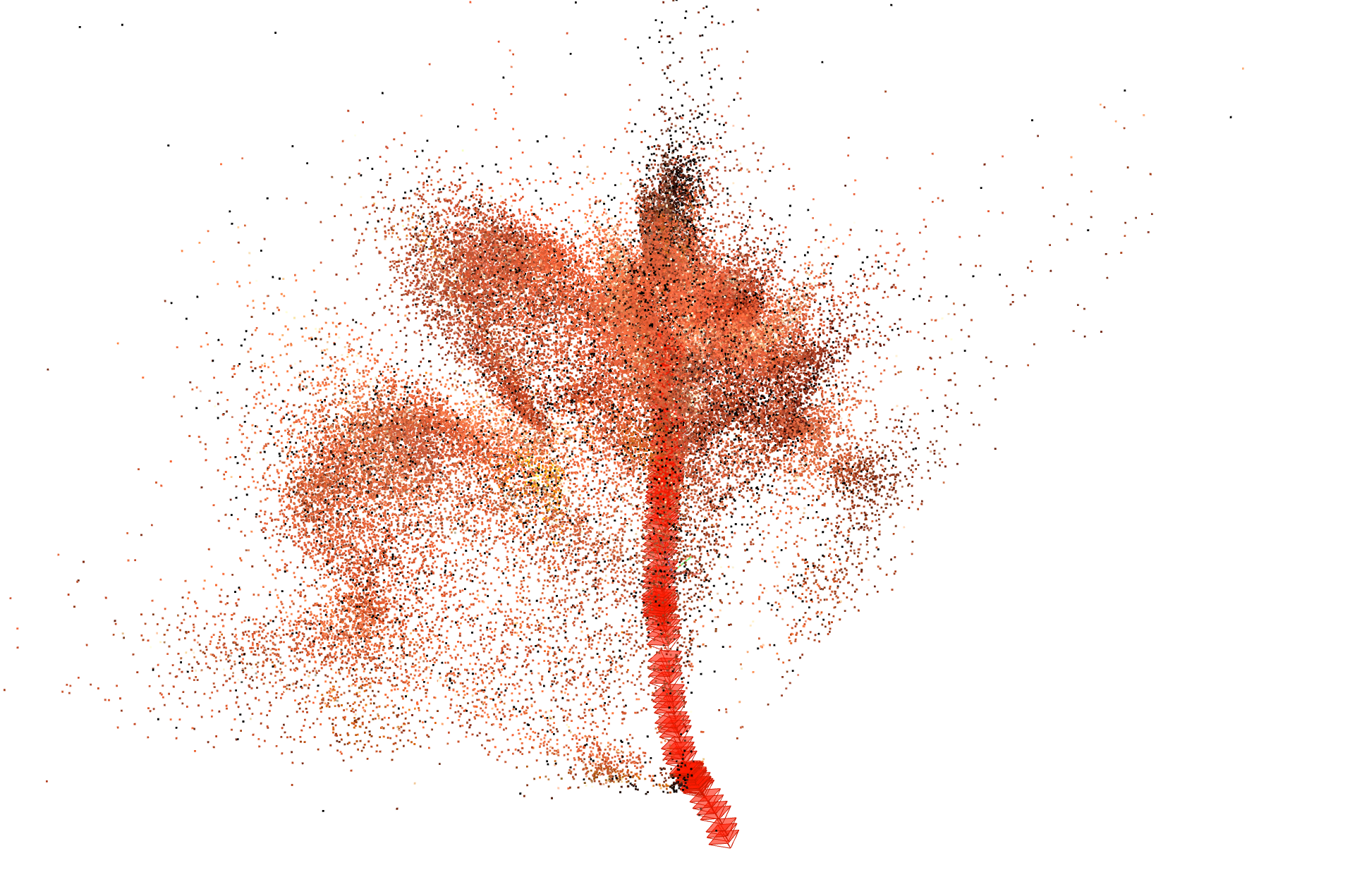} \\
         \end{tabular}
         \end{minipage}\\
         Subsequence Seq\_017\_1\\
         \begin{minipage}{0.42\linewidth}
         \begin{tabular}{@{}c@{}}
         \includegraphics[width=\linewidth]{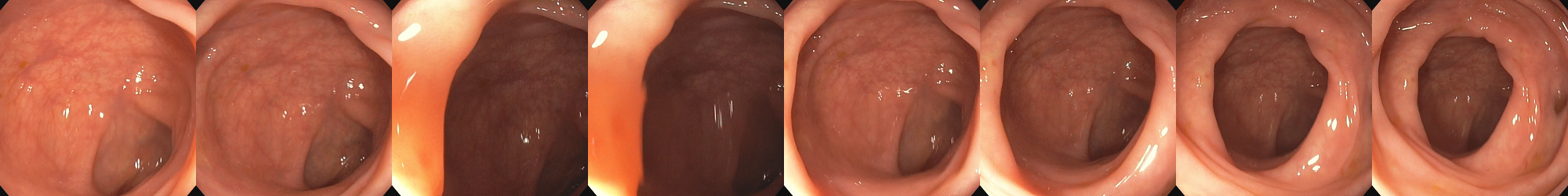} \\
         \includegraphics[width=\linewidth]{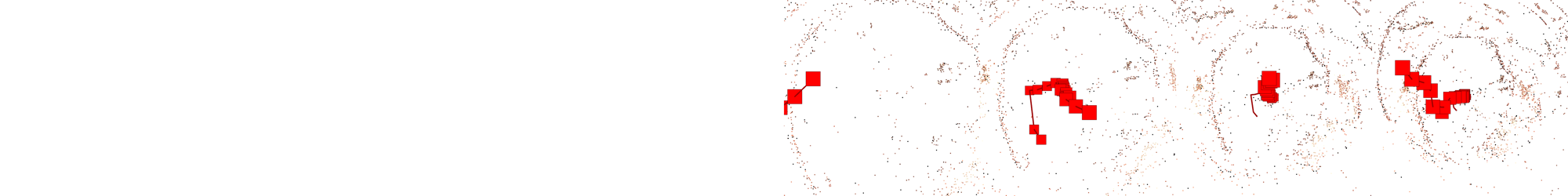} \\
         \includegraphics[width=\linewidth]{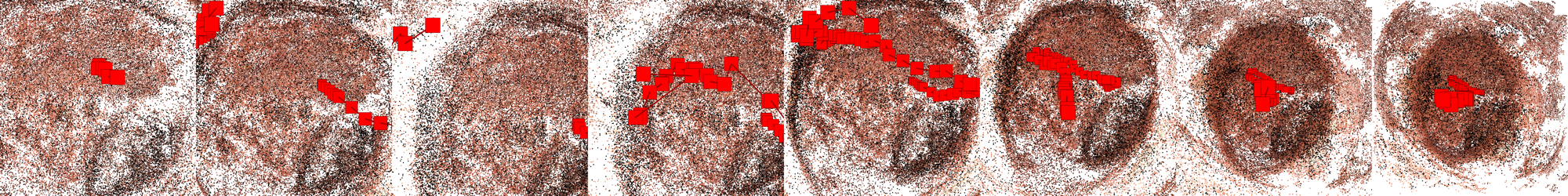} \\
         \end{tabular}
         \end{minipage}
         \begin{minipage}{0.42\linewidth}
         \begin{tabular}{@{}cc@{}}
         \includegraphics[width=0.48\linewidth,trim=10cm 6cm 20cm 10cm,clip]{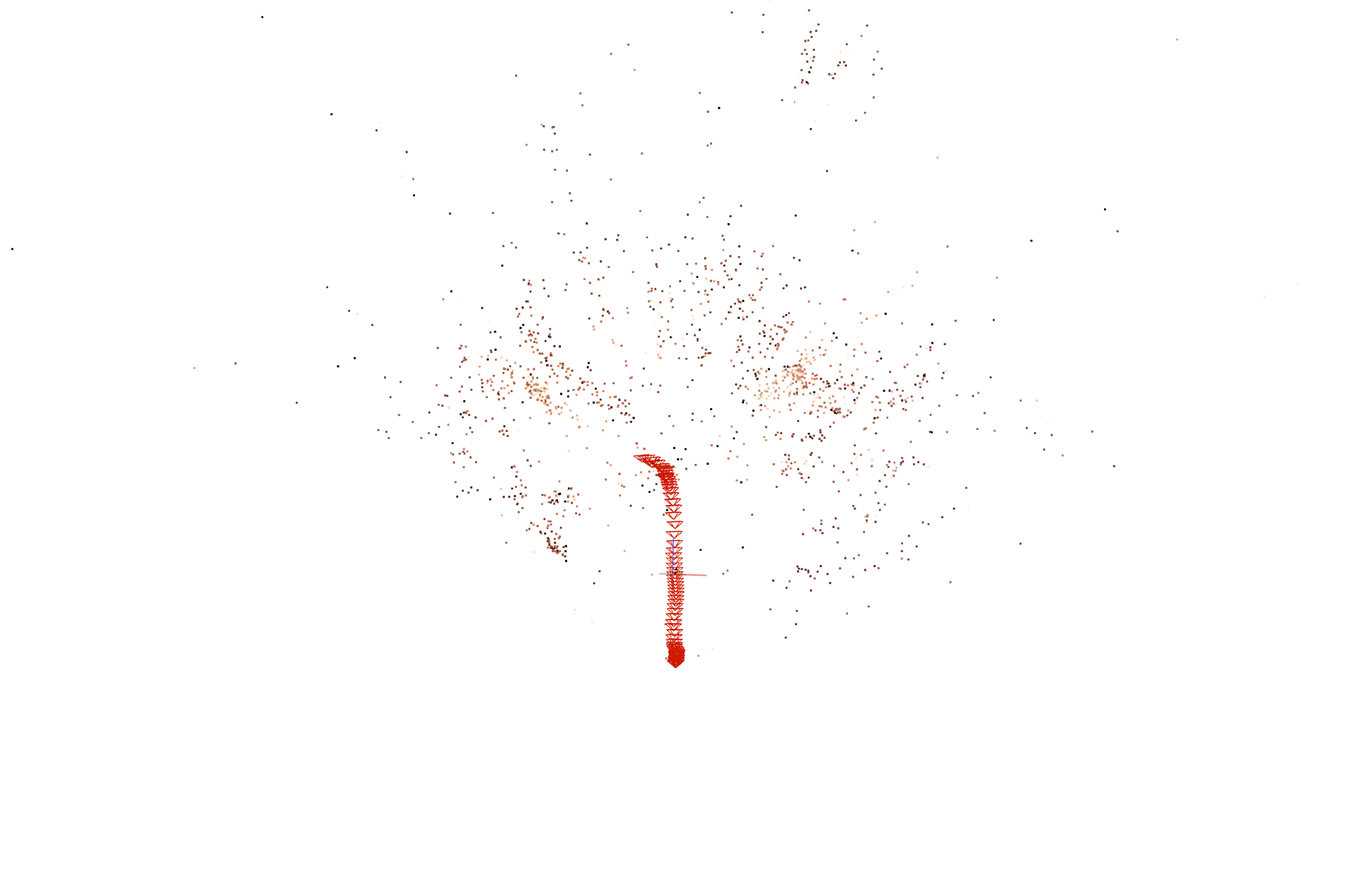} &
         \includegraphics[width=0.48\linewidth,trim=10cm 6cm 20cm 10cm,clip]{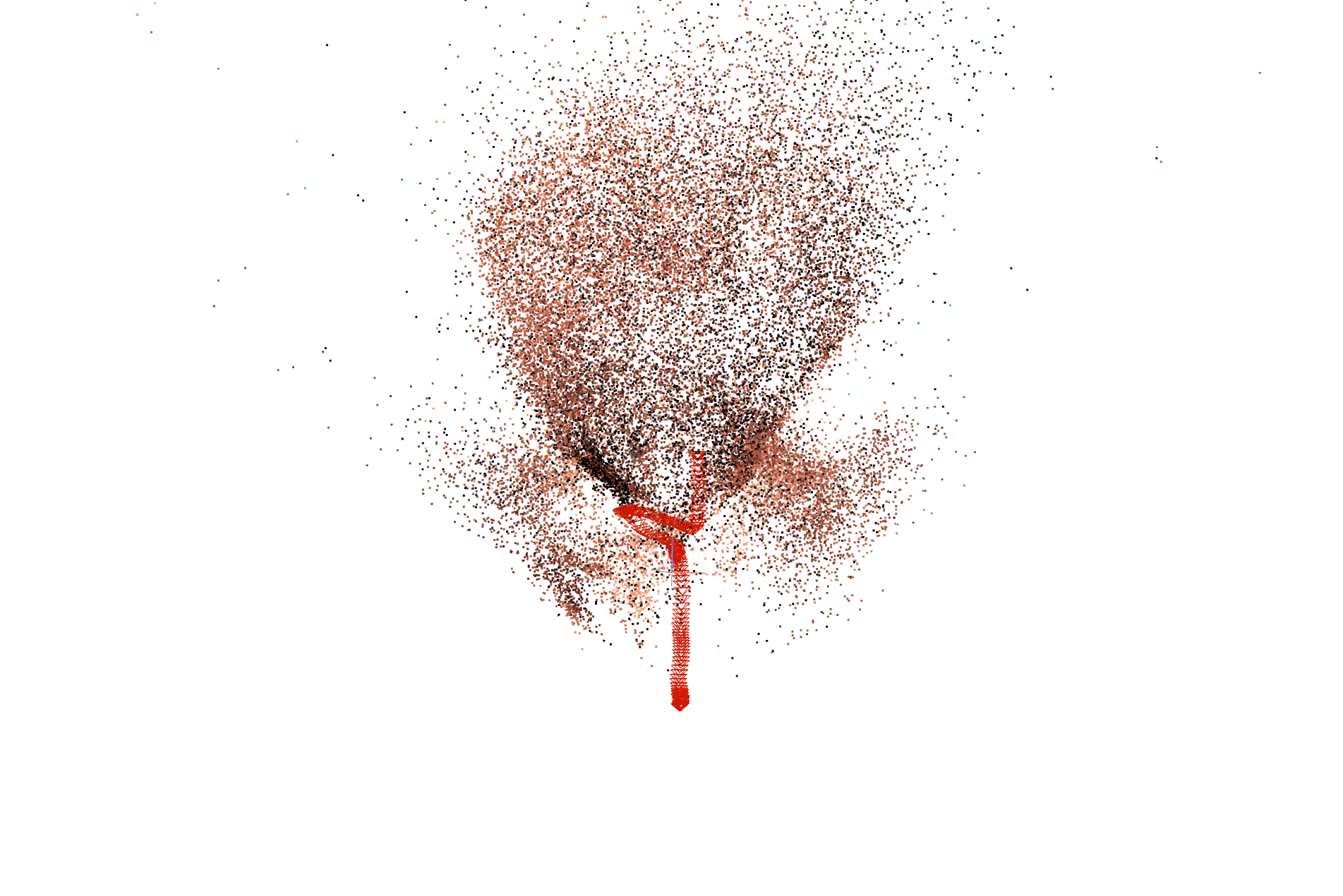} \\
         \end{tabular}
         \end{minipage}\\
         Subsequence Seq\_095\_1\\
         \begin{minipage}{0.42\linewidth}
         \begin{tabular}{@{}c@{}}
         \includegraphics[width=\linewidth]{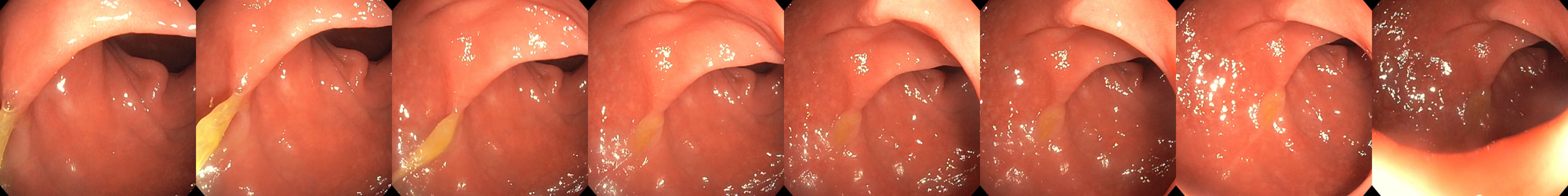} \\
         \includegraphics[width=\linewidth]{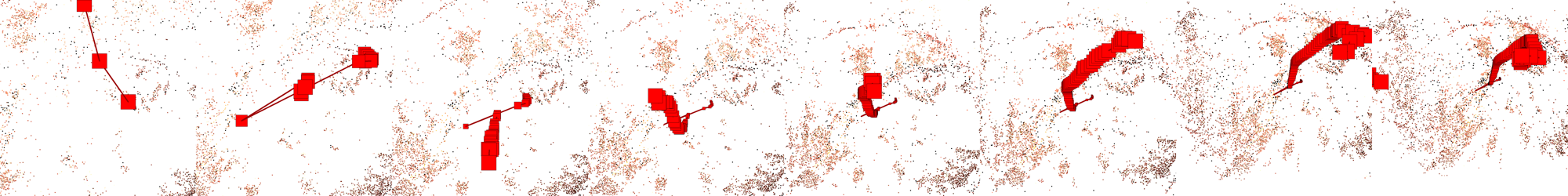} \\
         \includegraphics[width=\linewidth]{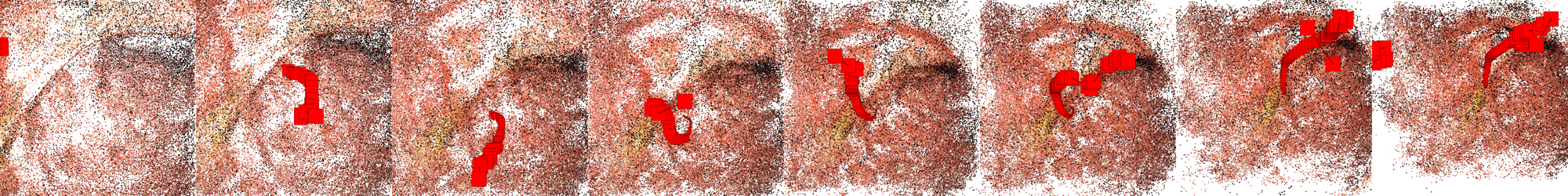} \\
         \end{tabular}
         \end{minipage}
         \begin{minipage}{0.42\linewidth}
         \begin{tabular}{@{}cc@{}}
         \includegraphics[width=0.48\linewidth,trim=10cm 10cm 20cm 5cm,clip]{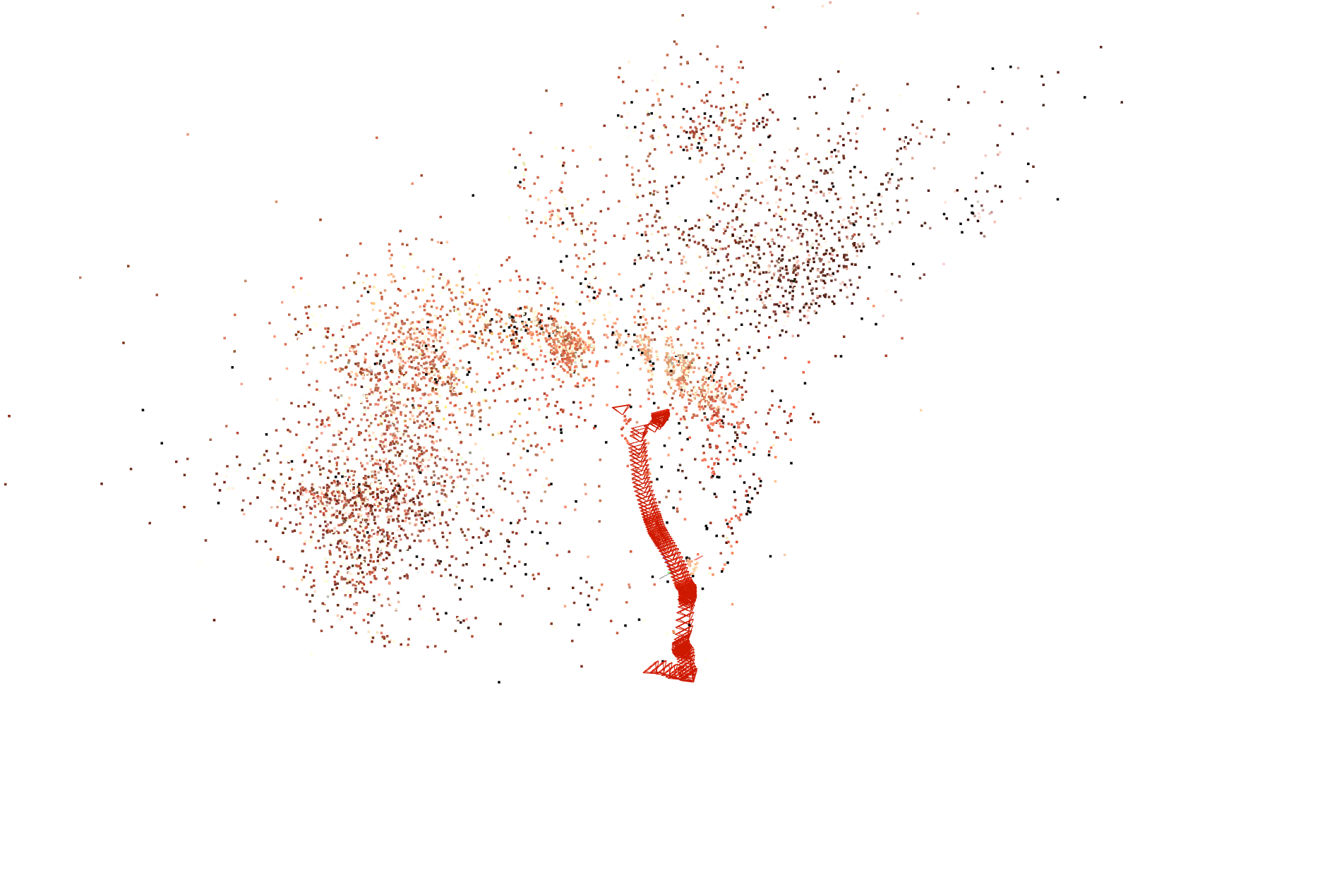} &
         \includegraphics[width=0.48\linewidth,trim=10cm 10cm 20cm 5cm,clip]{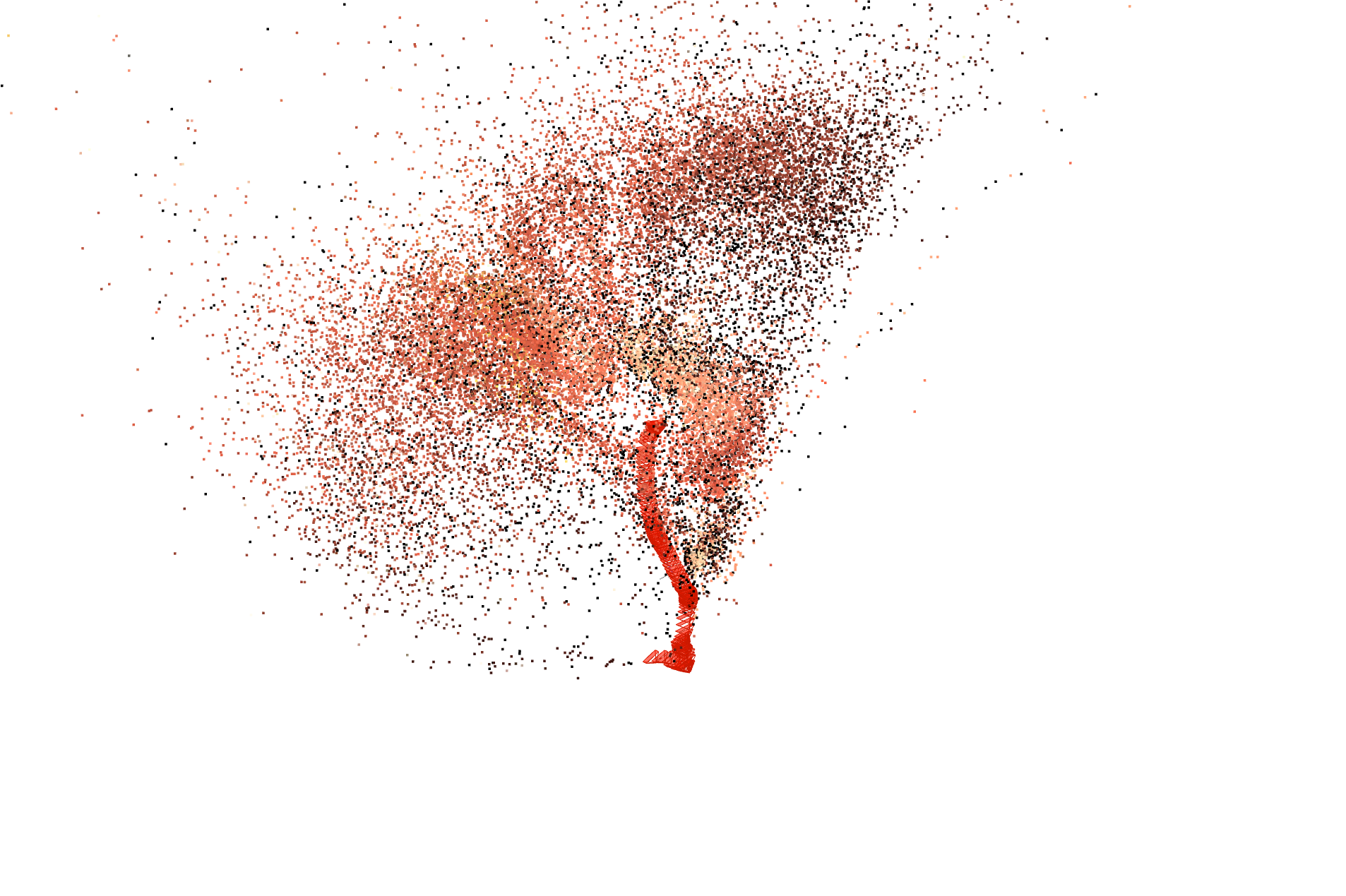} \\
        \end{tabular}
         \end{minipage}\\
    \end{tabular}
    \caption{Point clouds and camera trajectory reconstructed in 3 subsequences. Within each block, 
\textit{Left}: original frames aligned with reconstructed points. Top row contains 
eight evenly-spaced input sequence frames, and middle and bottom rows show 
corresponding reconstructed camera views of the point cloud using SIFT+GM and \mbox{SuperPoint-E}+GM respectively. \textit{Right:} SIFT and SuperPoint-E 3D reconstructions, including the camera poses (red markers).} 
    \label{fig:quality_examples}
\end{figure}

\noindent
Fig.~\ref{fig:quality_examples} (related animation available in the associated supplementary video) shows examples of the reconstructions in this experiment. The first example showcases the difference in density of the reconstruction. The second example shows how SP-E is also robust to abrupt illumination changes. In the third example, the camera trajectory is more stable and closer to real movement (first three frames in the left mosaics).

\noindent
Evaluating camera motion estimation in the \textit{C3VD-Test} set, with camera trajectory groundtruth available, the average RMS ATE for SIFT and SP-E is almost identical ($4.61$mm and $4.66$mm, respectively) but SP-E offers the benefits 
demonstrated above. 
In summary, Experiment 1 demonstrates that SuperPoint-E allows  significantly better reconstructions, with higher discriminative and repetitive capabilities, than other feature extraction methods.

\subsubsection{Complete sequence reconstruction coverage} 
\label{subsec:exp_full}

This experiment explores which approach can reconstruct more parts of a video and obtain larger reconstructions. We process full video sequences from \textit{EM-Full} using SIFT and \mbox{SuperPoint-E} and compare the video percentage that COLMAP reconstructs. 
The metrics used to analyze each sequence results are:
\begin{itemize}
    \item \textbf{Reconstructed}: amount of images included in at least one reconstruction.  
    \item \textbf{Reconstructed \%}: percentage of images included in at least one reconstruction. 
    \item \textbf{Reconstructions}: total number of reconstructions (or submaps) created. 
    \item \textbf{Average size}: number of images per reconstruction. 
\end{itemize}

\begin{figure*}[!tb]
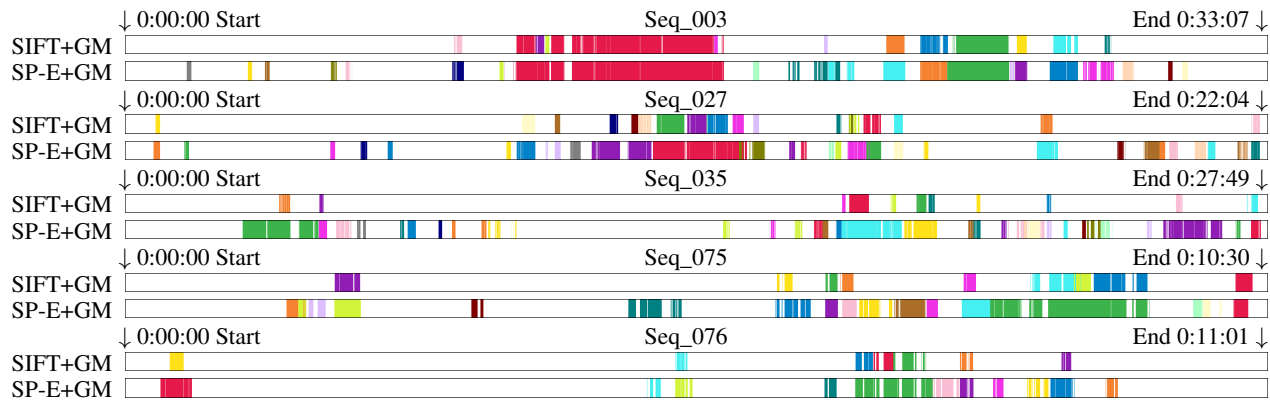

    \footnotesize
    \begin{tabularx}{\linewidth}{@{\;}c@{\;}l>{\centering\arraybackslash}Xr@{}}
         & $\downarrow$ 0:00:00 Start & Seq\_003 & End 0:33:07 $\downarrow$ \\
         SIFT+GM & \multicolumn{3}{@{\;}l@{}}{\includegraphics[width=0.92\linewidth]{images/colmap_bar_Seq_003_hd_sift_gm_25.png}}\\
         SP-E+GM & \multicolumn{3}{@{\;}l@{}}{\includegraphics[width=0.92\linewidth]{images/colmap_bar_Seq_003_hd_sptracking3_gm_25.png}}\\
         & $\downarrow$ 0:00:00 Start & Seq\_027 & End 0:22:04 $\downarrow$ \\
         SIFT+GM & \multicolumn{3}{@{\;}l@{}}{\includegraphics[width=0.92\linewidth]{images/colmap_bar_Seq_027_hd_sift_gm_25.png}}\\
         SP-E+GM & \multicolumn{3}{@{\;}l@{}}{\includegraphics[width=0.92\linewidth]{images/colmap_bar_Seq_027_hd_sptracking3_gm_25.png}}\\
         & $\downarrow$ 0:00:00 Start & Seq\_035 & End 0:27:49 $\downarrow$ \\
         SIFT+GM & \multicolumn{3}{@{\;}l@{}}{\includegraphics[width=0.92\linewidth]{images/colmap_bar_Seq_035_hd_sift_gm_25.png}}\\
         SP-E+GM & \multicolumn{3}{@{\;}l@{}}{\includegraphics[width=0.92\linewidth]{images/colmap_bar_Seq_035_hd_sptracking3_gm_25.png}}\\
         & $\downarrow$ 0:00:00 Start & Seq\_075 & End 0:10:30 $\downarrow$ \\
         SIFT+GM & \multicolumn{3}{@{\;}l@{}}{\includegraphics[width=0.92\linewidth]{images/colmap_bar_Seq_075_hd_sift_gm_25.png}}\\
         SP-E+GM & \multicolumn{3}{@{\;}l@{}}{\includegraphics[width=0.92\linewidth]{images/colmap_bar_Seq_075_hd_sptracking3_gm_25.png}}\\
         & $\downarrow$ 0:00:00 Start & Seq\_076 & End 0:11:01 $\downarrow$ \\
         SIFT+GM & \multicolumn{3}{@{\;}l@{}}{\includegraphics[width=0.92\linewidth]{images/colmap_bar_Seq_076_hd_sift_gm_25.png}}\\
         SP-E+GM & \multicolumn{3}{@{\;}l@{}}{\includegraphics[width=0.92\linewidth]{images/colmap_bar_Seq_076_hd_sptracking3_gm_25.png}}\\
    \end{tabularx}
    \caption{Coverage and reconstructed sections of full endoscopy videos based on SIFT or SP-E. Bar: full video timeline, left to right. Colored blocks:  frames included in the  reconstructions (the larger the colored parts, the more frames COLMAP is able to reconstruct given the chosen features). Same color means same reconstruction (or ``model'').}
    \label{fig:full_examples}
\end{figure*}

\noindent
Table~\ref{tab:exp_full} compares SIFT+GM and our SP-E model, summarizing this experiment. Note a clear improvement in coverage with SP-E, with double (\textit{Reconstructed \%} =$33.2\%$). Fig.~\ref{fig:full_examples} also illustrates this improvement.  
SP-E covers a much larger portion of the video, and the blocks are much longer. 
This second experiment shows that \mbox{SuperPoint-E} is more representative and robust in our target domain, obtaining more comprehensive models  of 
complete endoscopic sequences.

\begin{table}[!tb]
    \centering
    \footnotesize
    \caption{Reconstruction coverage metrics (average of the 5 sequences in \textit{EM-Full})}
    \begin{tabular}{l|c|c}
        \hline
         & SIFT+GM & SP-E+GM (Ours) \\ 
        \hline
        \textbf{Reconstructed} (images) & $934.8$ & $\textbf{2051.0}$ \\
        \textbf{Reconstructed \%} (\% images) & $15.1$\% & $\textbf{33.2}$\% \\
        \hline
        \textbf{Reconstructions} & $11.6$ & $\textbf{22.2}$ \\
        \textbf{Average size} (images) & $77.8$ & $\textbf{89.9}$ \\
        \hline
    \end{tabular}
    \label{tab:exp_full}
\end{table}

\subsubsection{Other endoscopic data} 
\label{subsec:exp_gastro}

\begin{figure*}[!tb]
    \footnotesize
    \centering
    \begin{tabular}{c@{}}
        \begin{minipage}{0.2\linewidth}
        (a-I)
        \end{minipage}
        \begin{minipage}{0.70\linewidth}
        \begin{tabular}{@{}c}
        \includegraphics[angle=90,width=0.70\linewidth]{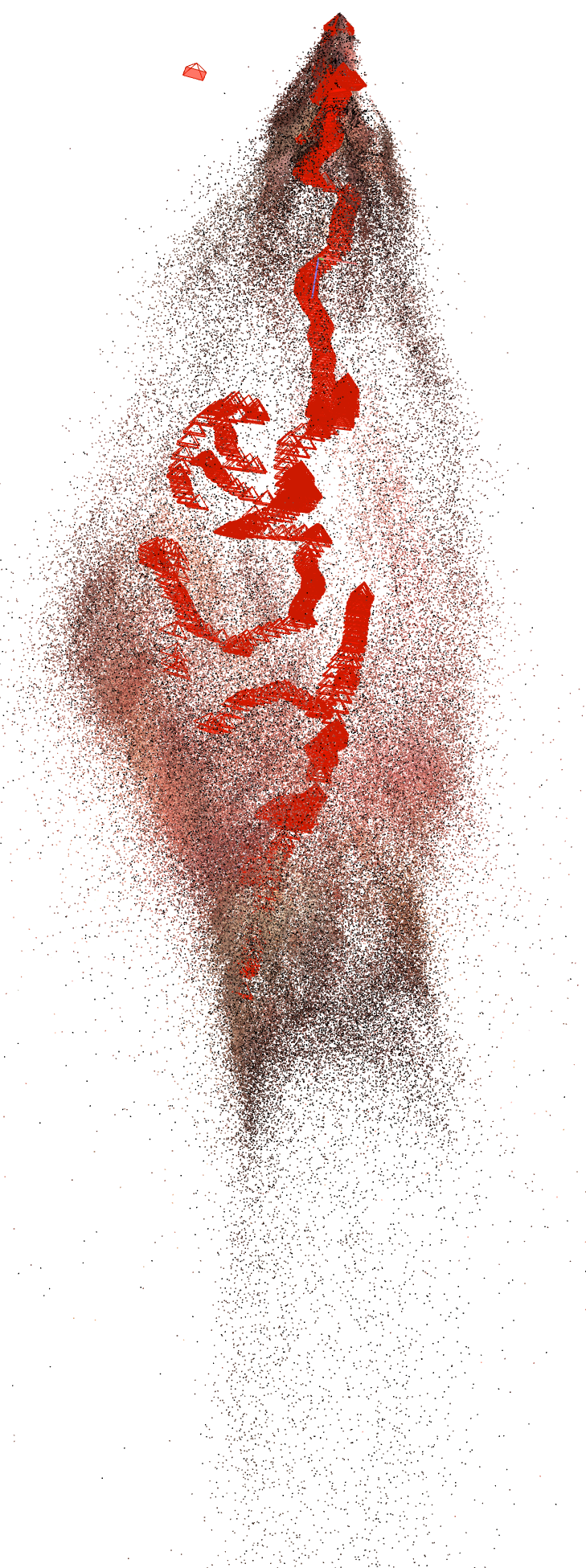}
        \end{tabular}
        \end{minipage}
    \end{tabular}
    \begin{tabular}{cccc}
        \includegraphics[width=0.20\linewidth]{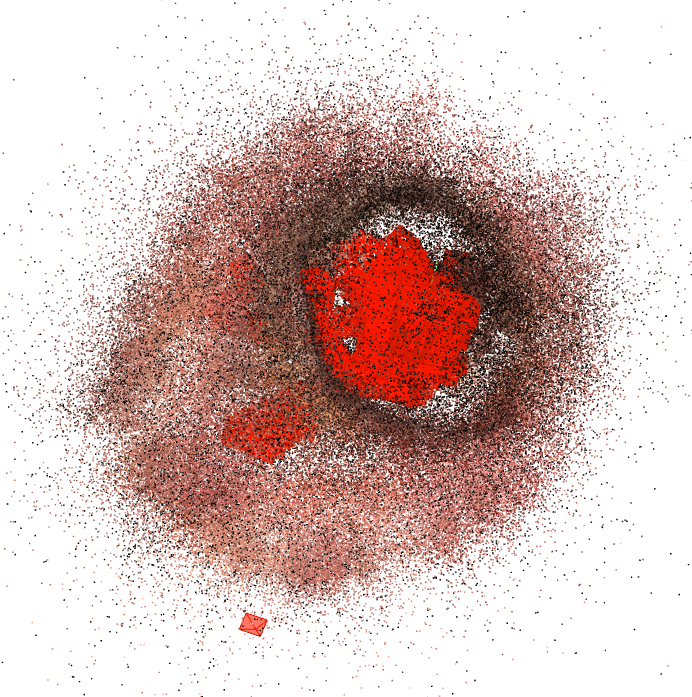} &
        \includegraphics[angle=90,width=0.20\linewidth]{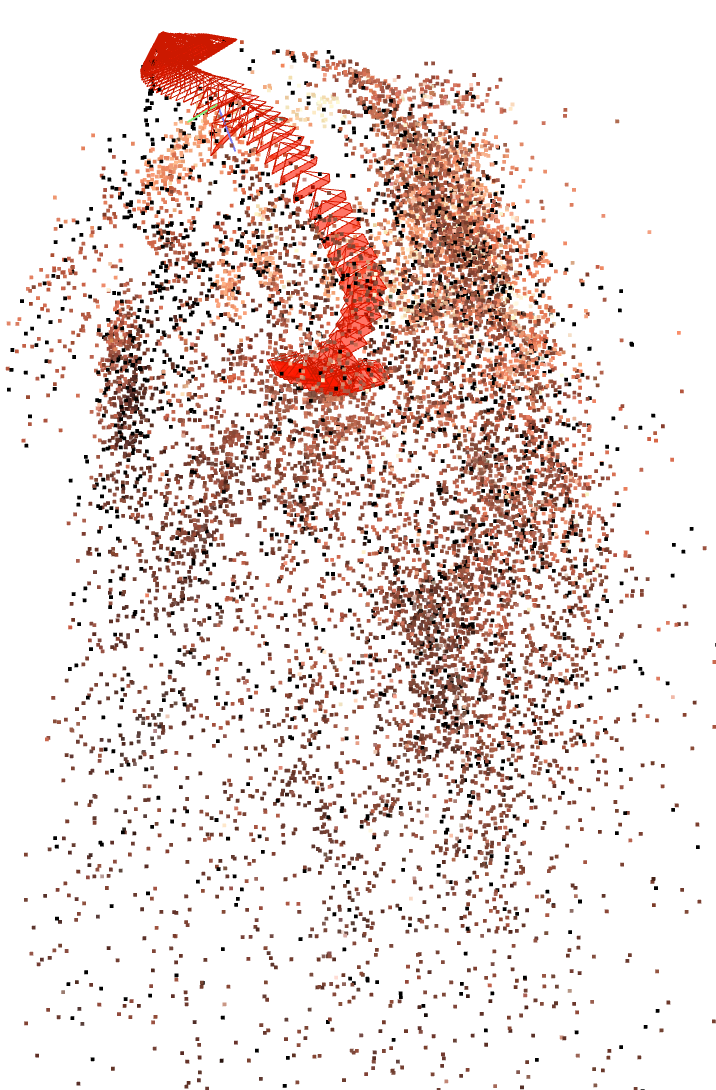} &
        \includegraphics[angle=90,width=0.20\linewidth]{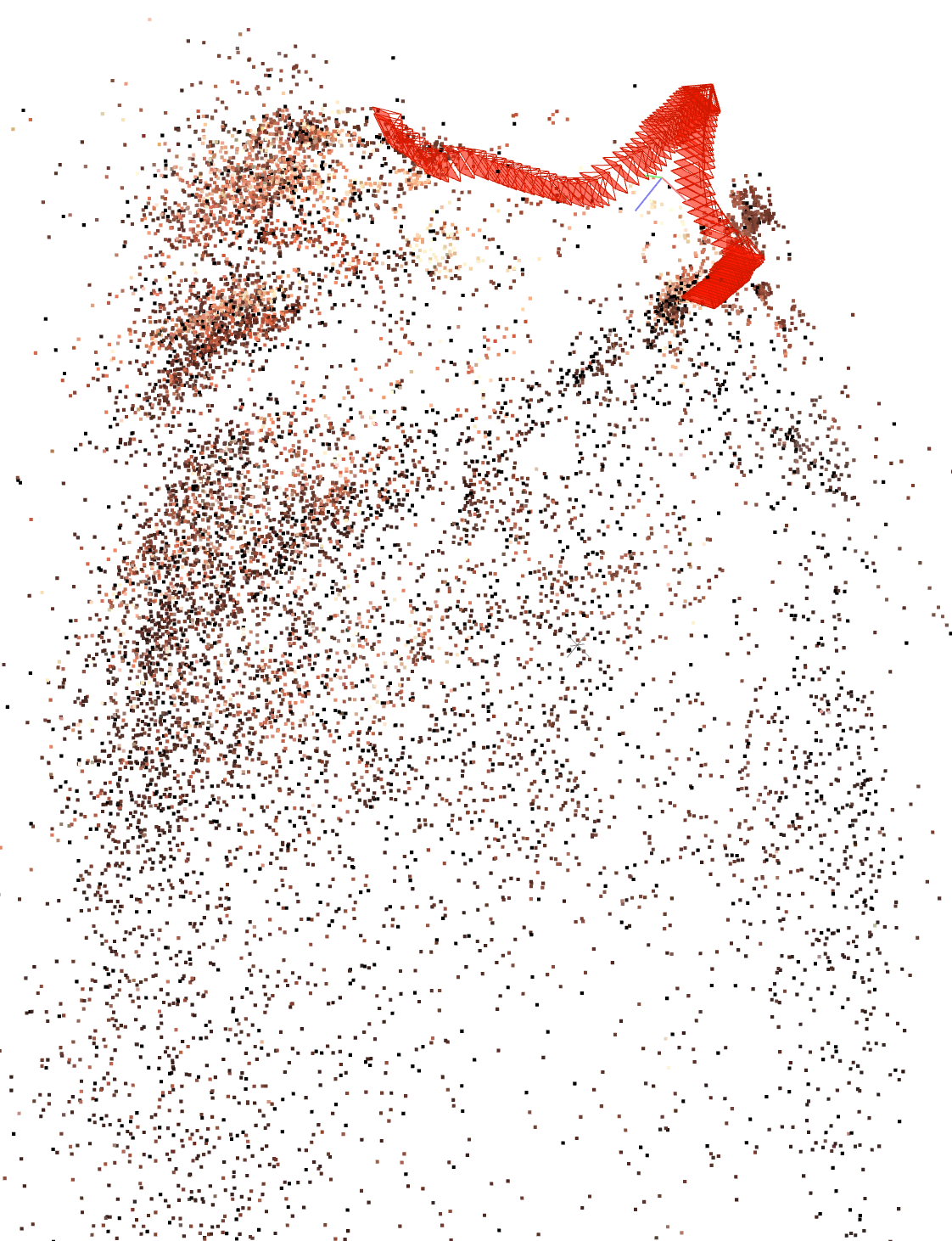} &
        \includegraphics[angle=90,width=0.20\linewidth]{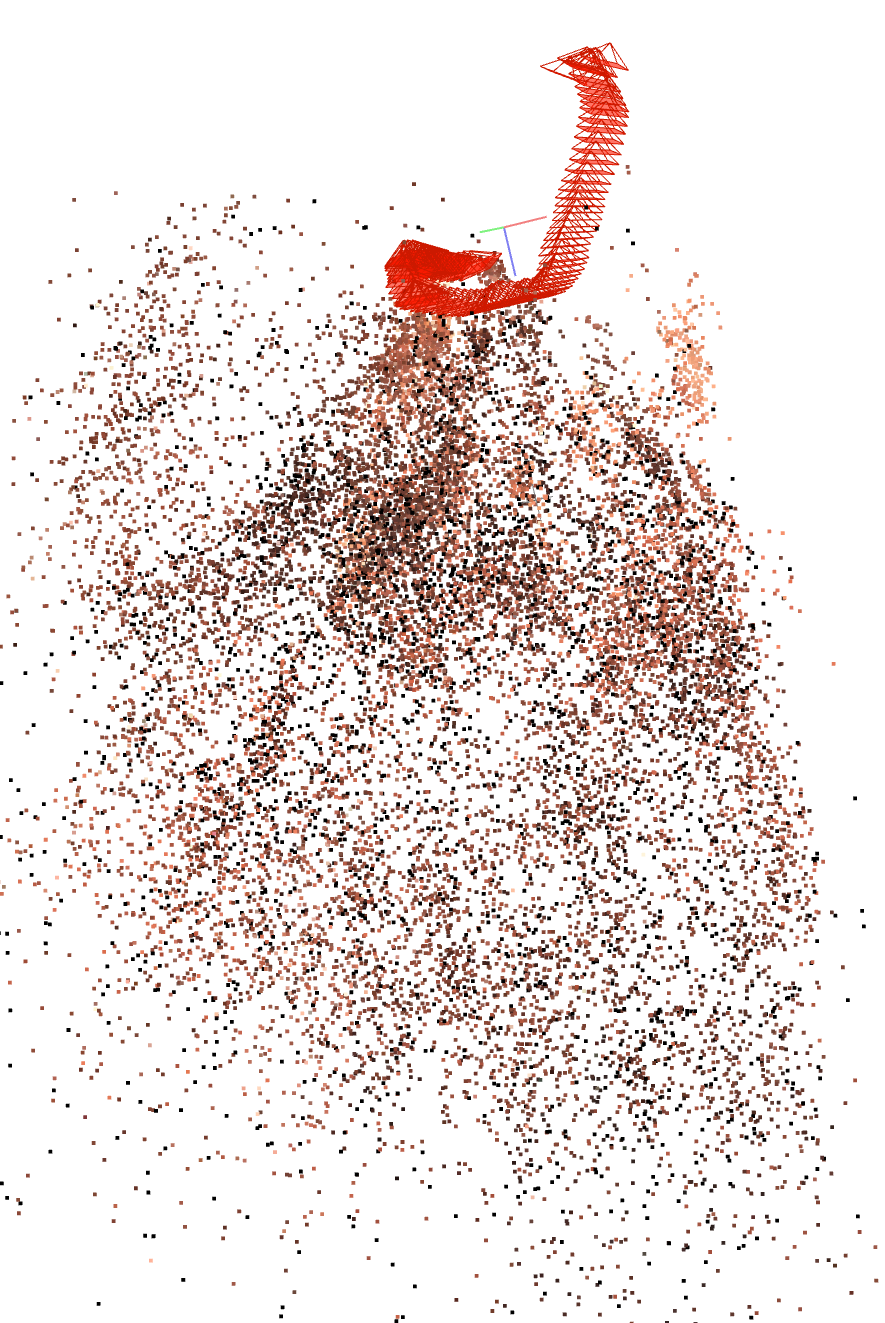} \\
        (a-II) & (b) & (c) & (d) \\
    \end{tabular}
    \caption{Examples of reconstructions on non-colonoscopy, endoscopy sequences. (a-I) and (a-II) are two views of an in-house bronchoscopy sequence while (b)-(d) are gastroscopy sequences from \textit{EM-Gastro}.} 
    \label{fig:gastro_examples}
\end{figure*}

While most work in this paper is done with colonoscopy data, the benefits of our method present potential for other endoscopy modalities. We have run experiments with real bronchoscopy and gastroscopy sequences, with promising results and similar  reconstruction quality as obtained in colonoscopy. 
Fig.~\ref{fig:gastro_examples} shows reconstructions obtained by our method in different domains  related to colonoscopy: a 1-minute-long \textbf{bronchoscopy} sequence is fully reconstructed, and the \textbf{gastroscopy} sequences from \textit{EM-Gastro} result in similar reconstructions as the colonoscopy results analyzed in the first experiment. Table~\ref{tab:exp_gastro} summarizes the quality of the gastroscopy (\textit{EM-Gastro} reconstructions. We observe that our proposed model performs much better also in this domain shift. Note that \mbox{SuperPoint-E} is not trained on gastroscopy sequences. 

\noindent
The conclusions from the experiment on colonoscopy data are also applicable here: more precise detection, more images reconstructed, and more reconstructed points with longer average tracks. The features are also better spread over the images with a very low percentage of points on top of specularities. This experiment suggests that \mbox{SuperPoint-E}'s benefits for feature extraction look promising beyond colonoscopy.

\begin{table}[!tb]
    \centering
    \footnotesize
    \caption{Reconstruction quality in gastrocopy susequences. All metrics are the average over the 5 sequences in \textit{EM-Gastro}.}
    \begin{tabular}{l|c|c}
        \hline
         & SIFT+GM & SP-E+GM (Ours) \\ 
        \hline
        \textbf{Precision} & $40.9$\% & $\textbf{53.1}$\% \\
        \textbf{Reconstructed} & $80.2$\% & $\textbf{99.8}$\% \\
        \textbf{3D points} & $5$K & $\textbf{27}$K \\
        \textbf{Track length} & $7.02$ & $\textbf{8.19}$ \\
        \textbf{MAE} & $\textbf{1.36}$ & $1.71$ \\
        \textbf{MAE of 10K} & $1.36$ & $\textbf{1.10}$ \\
        \textbf{Spread} & $36.7$\% & $\textbf{60.2}$\% \\
        \textbf{Specular} & $19.9$\% & $\textbf{4.5}$\% \\
        \hline
    \end{tabular}
    \label{tab:exp_gastro}
\end{table}%

\subsubsection{Model analysis and tuning} 
\label{sec:tuning}

Table~\ref{tab:exp_batch} summarizes the analysis of different alternatives for the core design choices in our method. 
Second and third columns 
analyze the impact of training  only with SIFT features as source for the 3D reconstructions (src=SIFT) vs. using reconstructions from both SIFT and SuperPoint features (src=SIFT+SP). The latter option is clearly superior, thus being our choice. 
Fourth, fifth and sixth columns
show the results with different $N$ values: $N$=$2$, for a training batch of only a pair of images, chosen randomly, that share at least one reliable track; $N$=$4$, a training batch of four images chosen the same way; $N\in[4,12]$, a training batch of variable size between 4 and 12, we cannot choose a high number strictly, e.g. $N$=$12$, because almost half of the reliable tracks are shorter (44\%). Note improvements in \textit{Precision}, \textit{3D points} and \textit{Track length} using $N$=$4$, but including larger batches ($N\in[4,12]$) does not significantly improve. Then, we set $N$=$4$ in our approach. 
Seventh and eighth columns
show the impact of changing SP-E's \texttt{\small keypoint\_threshold}, from 0.015 to 0.0005. The lower this SP threshold, the more points are accepted. Our study suggests that they are all quite relevant, 
suggesting that SP-E features are very well suited for SfM in endoscopy, even without filtering them with a high detection threshold. 

\begin{table*}[!tb]
    \centering
    \footnotesize
    \caption{Reconstruction quality metrics for \textbf{SP-E configurations} using BF matching. Header row: fixed parameters and parameters analyzed in each block. We consider different supervision sources src (SIFT or SIFT+SP) and different values of $N$ and \texttt{\small keypoint\_threshold}(th) in training. All metrics are the average over the 7 sequences in \textit{EM-Test}.}
    \begin{tabular}{@{}l|c|c||c|c|c||c|c}
        \hline
        \multicolumn{1}{r|}{fixed$\rightarrow$} & \multicolumn{2}{c||}{$N$=$2$, th=0.015} & \multicolumn{3}{c||}{src=SIFT+SP, th=0.015} & \multicolumn{2}{c}{src=SIFT+SP, $N$=$4$} \\
        \multicolumn{1}{r|}{analyzed$\rightarrow$} & src=SIFT & SIFT+SP & $N$=$2$ & =$4$ & $\in[4,12]$ & th=0.015 & =0.0005 \\ 
        \hline
        \textbf{Precision} & $55.5$\% & $\textbf{57.7}$\% & $57.7$\% & $\textbf{59.6}$\% & $58.5$\% & $59.6$\% & $\textbf{60.5}$\% \\
        \textbf{Reconstructed} & $\textbf{99.1}$\% & $\textbf{99.1}$\% & $\textbf{99.1}$\% & $\textbf{99.1}$\% & $98.95$\% & $99.1$\% & $\textbf{99.5}$\% \\
        \textbf{3D points} & $13$K & $\textbf{28}$K & $28$K & $34$K & $\textbf{35}$K & $34$K & $\textbf{76}$K \\
        \textbf{Track length} & $9.01$ & $\textbf{9.45}$ & $9.45$ & $\textbf{9.53}$ & $\textbf{9.53}$ & $9.53$ & $\textbf{10.78}$ \\
        \textbf{MAE} & $\textbf{1.69}$ & $1.74$ & $\textbf{1.74}$ & $1.75$ & $1.76$ & $\textbf{1.75}$ & $1.78$ \\
        \textbf{MAE of 10K} & $1.51$ & $\textbf{1.13}$ & $1.13$ & $1.02$ & $\textbf{1.01}$ & $1.02$ & $\textbf{0.69}$ \\
        \textbf{Spread} & $46.4$\% & $\textbf{60.2}$\% & $60.2$\% & $67.5$\% & $\textbf{67.6}$\% & $67.5$\% & $\textbf{85.2}$\% \\
        \textbf{Specular} & $21.6$\% & $\textbf{12.4}$\% & $12.4$\% & $\textbf{10.0}$\% & $11.0$\% & $10.0$\% & $\textbf{6.2}$\% \\
        \hline
    \end{tabular}
    \label{tab:exp_batch}
\end{table*}

\noindent
Table~\ref{tab:exp_time} provides a breakdown of the processing time in the SfM pipeline. Both SIFT and SuperPoint-E used GPU acceleration. The two matching algorithms did not use GPU acceleration.   
Times from one 136-frames-long subsequence from sequence 1 in \textit{EM-Test}. 
Computer setup: 11th Gen Intel® Core™ i7-11700K @ 3.60GHz × 16 CPU, 64GB RAM and a NVIDIA GeForce RTX 3090 24GB GPU. 

\begin{table}[!tb]
    \centering
    \footnotesize
    \caption{Processing time for feature extraction and matching measured in one subsequence (from seq. 1 in \textit{EM-Test}).  
    }
    \begin{tabular}{@{}l@{}|c|c}
        \hline
         & SIFT & SP-E \\ 
        \hline
        Image size (pixels) & \multicolumn{2}{c}{1440$\times$1080} \\
        \hline
        Descriptor size (\texttt{\small i}=integer, \texttt{\small f}=floating point number) & $128$\texttt{\small i} & $256$\texttt{\small f} \\
        \hline
        \hline
        \textbf{Detection (per image, GPU) time in ms}  & $31$ & $\textbf{20}$ \\
        * Detected points (per image) & $7$K & $\textbf{10}$K \\
        \hline
        \hline
        \textbf{Brute Force (per image pair, no GPU) time in ms}  & $\textbf{57}$ & $290$ \\ 
        * Points included in final 3D reconstruction & $31$K & $\textbf{71}$K \\
        \hline
        \textbf{Guided Match. (per image pair, no GPU)  time in ms} & $\textbf{236}$ & $1\,282$ \\ 
        * Points included in final 3D reconstruction & $46$K & $\textbf{76}$K \\
        \hline
        \multicolumn{3}{@{}p{8.5cm}}{* \textit{To illustrate the trade-off between time and reconstruction quality, we provide amount of detected and reconstructed points.}}
    \end{tabular}
    \label{tab:exp_time}
\end{table}

\noindent
\mbox{SuperPoint-E} uses 256-long floating point descriptor vectors, which means an higher overall processing time than SIFT, that uses 128-long integers on its COLMAP implementation. 
The most significant execution time overload occurs in the Guided Matching step, as expected. Noticeably, our previous experiment also highlighted how the Guided Matching strategy is not critical for our model to generate better 3D reconstructions (denser and covering larger portions of the scene) than the other baselines. The 3D points 
metric shows that our method using Brute Force matching obtains a reconstruction of double the amount of 3D points that SIFT + Guided Matching does, with similar processing time requirements. For resource constrained application, \mbox{SuperPoint-E} opens the possibility to omit the Guided Matching step while obtaining much denser 3D reconstructions than other methods.

\section{Conclusions}
Our new feature extraction method, SP-E, improves the performance of SfM in endoscopy videos. 
Our main contribution within this method, \textit{Tracking Adaptation}, is a novel training strategy for feature extraction models like SuperPoint. This strategy focuses on learning how to extract features that are ``reliable'', i.e., features that are re-extracted with discriminative descriptors across multiple views of the same scene. 
Our experiments demonstrate how \mbox{SuperPoint-E} extracts repeatable features with discriminative descriptors, well spread over the image and avoiding specular reflections. For SfM in endoscopy, this results in denser and longer reconstructions. 
We have adapted the Guided Matching implementation in COLMAP so it can be used with our features, enforcing a globally consistent model on the matches. This brings only slightly better results than a regular Brute Force matching, which only relies on the descriptors to find unique matches. This further proves that the features extracted by \mbox{SuperPoint-E} are much more reliable and discriminative for endoscopy videos. 
Our model's features are also shown to  expand the size and the number of submaps that COLMAP detects when processing complete real colonoscopies.

\subsection*{Acknowledgements} 
This project has been funded by the European Union’s Horizon 2020 research and innovation programme under grant agreement No 863146 and Arag\'{o}n Government project  T45\_23R.

\bibliographystyle{splncs04}
\bibliography{biblio.bib}

\end{document}